\documentclass[11pt,letterpaper]{article}
\usepackage[a4paper,bindingoffset=0.2in,%
left=0.8in,right=0.8in,top=1in,bottom=1in,%
footskip=.25in]{geometry}
\usepackage{graphicx}
\usepackage{booktabs}
\usepackage{url}
\usepackage{cite}
\usepackage[title]{appendix}
\usepackage[utf8]{inputenc} 
\usepackage[T1]{fontenc}    
\usepackage{hyperref}       
\usepackage{url}            
\usepackage{booktabs}       
\usepackage{amsfonts}       
\usepackage{nicefrac}       
\usepackage{microtype}      

\usepackage{algorithm}
\usepackage{algorithmic}
\usepackage{enumitem}
\usepackage{authblk}
\usepackage{breakcites}
\usepackage{amsmath}
\usepackage{bbm}
\usepackage{verbatim}
\usepackage[font=small]{caption}
\usepackage{amsthm}
\usepackage{subcaption}
\usepackage{mathtools}
\usepackage{xcolor}

\usepackage{graphicx}
\usepackage{booktabs}
\usepackage{url}
\usepackage{textcomp}
\usepackage{hyperref}
\usepackage{nicefrac}
\usepackage{microtype}
\usepackage{algorithm}
\usepackage{etoolbox}
\usepackage{soul}
\usepackage{ulem}
\let\classAND\AND
\let\AND\relax
\usepackage{algorithmic}

\let\AND\classAND
\AtBeginEnvironment{algorithmic}{\let\AND\algoAND}
\usepackage{breakcites}	
\usepackage{amsfonts,amsmath,bm,bbm,amsthm, amssymb}
\usepackage{verbatim}
\usepackage{caption}
\usepackage{subcaption}
\usepackage{mathtools}
\usepackage{anyfontsize}

\usepackage{amsmath,amsfonts,bm}

\def\eqref#1{equation~\ref{#1}}

\def\1{\bm{1}}

\DeclareMathAlphabet{\mathsfit}{\encodingdefault}{\sfdefault}{m}{sl}
\SetMathAlphabet{\mathsfit}{bold}{\encodingdefault}{\sfdefault}{bx}{n}

\DeclareMathOperator*{\argmax}{arg\,max}

\newtheorem{theorem}{\bf Theorem}
\newtheorem*{proposition*}{\bf Proposition}

\newtheorem{corollary}{\bf Corollary}[theorem]
\newtheorem{lemma}[theorem]{\bf Lemma}
\newtheorem{proposition}[theorem]{\bf Proposition}
\newtheorem{definition}{\bf Definition}
\newtheorem{remark}[theorem]{\bf Remark}
\newtheorem{claim}{\bf Claim}

\newcommand{\calx}{\mathcal{X}}
\newcommand{\calv}{\mathcal{V}}
\newcommand{\expec}{\mathbb{E}}
\newcommand{\prob}{\mathbb{P}}
\newcommand{\undr}{\underline{r}}
\newcommand{\overr}{\overline{r}}
\newcommand{\ind}{\mathbbm{1}}

 \newcommand{\ignore}[1]{} 

\newcommand\numberthis{\addtocounter{equation}{1}\tag{\theequation}}

\usepackage{centernot}

\newcounter{const-no}

\def\EE{{\mathbb{E}}}
\def\PP{{\mathbb{P}}}

\allowdisplaybreaks

\title{Bandits with Stochastic Experts: Constant Regret, Empirical
Experts and Episodes}

\author[1]{Nihal Sharma\thanks{E-mail: \texttt{nihal.sharma@utexas.edu}.}
}
\author[2]{Rajat Sen}
\author[2]{Soumya Basu}
\author[3]{Karthikeyan Shanmugam \thanks{This work was conducted when Karthikeyan Shanmugam was at IBM Research, New York, USA.}}
\author[1]{Sanjay Shakkottai}
\affil[1]{The University of Texas at Austin, Austin, USA}
\affil[2]{Google, Mountain View, USA}
\affil[3]{Google DeepMind, Bengaluru, India}

\begin{document}

\maketitle

\begin{abstract}
We study a variant of the contextual bandit problem where an agent can intervene through a set of stochastic expert policies. Given a fixed context, each expert samples actions from a fixed conditional distribution. The agent seeks to remain competitive with the `best' among the given set of experts. We propose the Divergence-based Upper Confidence Bound (D-UCB) algorithm that uses importance sampling to share information across experts and provide horizon-independent constant regret bounds that only scale linearly in the number of experts. We also provide the Empirical D-UCB (ED-UCB) algorithm that can function with only approximate knowledge of expert distributions. Further, we investigate the episodic setting where the agent interacts with an environment that changes over episodes. Each episode can have different context and reward distributions resulting in the best expert changing across episodes. We show that by bootstrapping from $\mathcal{O}\left(N\log\left(NT^2\sqrt{E}\right)\right)$ samples, ED-UCB guarantees a regret that scales as $\mathcal{O}\left(E(N+1) + \frac{N\sqrt{E}}{T^2}\right)$ for $N$ experts over $E$ episodes, each of length $T$. We finally empirically validate our findings through simulations.
\end{abstract}

\maketitle

\section{Introduction}
Recommendation systems for suggesting items to users are commonplace in online services such as marketplaces, content delivery platforms and ad placement systems. Such systems, over time, learn from user feedback, and improve their recommendations. An important caveat, however, is that both the distribution of user types and their respective preferences change over time, thus inducing changes in the optimal recommendation and requiring the system to periodically ``reset'' its learning.

We consider systems with known change-points (aka episodes) in the distribution of user-features and preferences. Examples include seasonality in product recommendations where there are marked changes in interests based on time-of-year, or ad-placements based on time-of-day. While a baseline strategy would be to re-learn the recommendation algorithm in each episode, it is often advantageous to share some learning across episodes. Specifically, one often has access to (potentially, a very) large number of pre-trained recommendation algorithms (aka experts), and the goal then is to quickly determine (in an online manner) which expert is best suited to a specific episode. Crucially, the expert policies are \textit{invariant over episodes} and can be learnt offline -- given samples of (observed context, recommended action) pairs from the deployment of one expert in the past, one can infer the policy \textit{approximately}. The problem is then to efficiently `transfer' this approximate knowledge to the online phase to accelerate the learning of the episode-dependent best expert.

As a specific example, we take the case of online advertising agencies which are companies that have proprietary ad-recommendation algorithms that place ads for other product companies on newspaper websites based on past \textit{campaigns}. In each campaign, the agencies places ads for a specific product of the client (e.g., a flagship car, gaming consoles, etc) in order to maximize the click-through rate of users on the newspaper website. At any given time, the agency signs contracts for \textit{new} campaigns with new companies. The information about product features and user profiles form the context, whose distribution changes across campaigns due to change in user traffic and updated product line ups. This could also cause shifts in user preferences. In practice, the agency already has a finite inventory of ad-recommendation models (aka experts, typically logistic models for their very low inference delays of micro-seconds that is mandated by real-time user traffic) from past campaigns. On a new campaign, online ad agencies \textit{bid} for slots in news media outlets depending on the profile of the user that visits their website, using these pre-learned experts (see \cite{perlich2014machine, zhang2014optimal}). In this setting, agencies only re-learn which experts in their inventory works best for their new campaign, possibly fine-tuning them across campaigns. Our work models this episodic setup, albeit, without fine tuning of experts between campaigns.

In this paper, we study the contextual bandit problem with stochastic experts in the episodic setting. In the single-episode case, we develop an Importance Sampling-based strategy that shares information across experts and provides horizon-independent regret guarantees when expert policies and context distributions are known. In the episodic case, we generalize our methods to function with approximate knowledge of these quantities. 

\subsection{Main contributions}

We formulate the Contextual Bandit with Stochastic Experts problem. Here, an agent interacts with an environment through a set of $N$ experts. Each expert $i$ is characterized by a fixed and unknown conditional distribution over actions in a set $\calv$ given the context from $\calx$. We also extend this to the episodic case, where this agent-environment interaction is carried out over $E$ episodes. At the start of episode $e$, the context distribution $p_e(\cdot)$ as well the distribution of rewards $q_e(\cdot|v,x)$ changes and remains fixed over the length of the episode denoted by $T$. At each time, the agent observes the context $X$, chooses one of the $N$ experts and plays the recommended action $V$ to receive a reward $Y$. Note here that the expert policies remain invariant across {\bf all} episodes. 

The goal of the agent is to track the best expert in order to maximize the cumulative sum of rewards. Here, the best expert is one that generates the maximum average reward averaged over the randomness in contexts, recommendations and rewards. In the episodic setting, the agent seeks to track the \textit{episode-dependent} best expert, which may change across episodes due to differences in the environment. Due to the stochastic nature of experts, we can use Importance Sampling (IS) estimators to share reward information to leverage the information leakage. 

Our main contributions are as follows:

\noindent{\bf 1. Divergence-Based Upper Confidence Bound (D-UCB) Algorithm:} We develop the D-UCB algorithm (Algorithm \ref{alg:DUCB}) for the contextual bandit with stochastic experts problem which employs a clipped IS-based estimator to predict expert rewards. We analyze this estimator and show exponentially fast concentrations around its mean in Theorem \ref{lem:clipped}. We also provide horizon-independent regret guarantees for D-UCB in Theorem \ref{thm: one episode regret} that scale as $\mathcal{O}(C_1 N)$ with $N$ experts where $C_1$ is a problem-dependent constant. Additionally, we also extend this to the case where the expert policies are only known approximately and provide the Empirical D-UCB (ED-UCB) algorithm (Algorithm \ref{alg:EDUCB}) for this setting. We also show that with well-approximated experts, using ED-UCB leads to regret performance that scales as $\mathcal{O}(C_2 N)$ with $C_2$ as another constant  (similar to that of the full-information setting of D-UCB) in Theorem \ref{thm: per-epi regret main paper}.

Further, in Section \ref{sec: improved scaling}, we also show that, with some mild assumptions, these regret bounds can be improved to $\mathcal{O}(\log N)$ and present strategies to improve computational complexity of our algorithms at the cost of some regret.

\noindent{\bf 2. Theoretical Contributions:} Authors in \cite{sen2017identifying} study the best-arm identification problem in our setting and design a successive elimination algorithm wherein the sequence of expert plays in each epoch is decided before any samples are observed. Thus, they provide concentration results for Clipped IS-based estimators in the setting where the number of samples collected under each expert is a priori known using Chernoff-type analyses. 

In contrast, we study the cumulative regret setting in this work and develop a Upper Confidence Bound-style \textit{randomized bandit algorithm} to choose experts at each time step. This results in the number of samples under each expert at any time being a random quantity, which restricts the use of Chernoff bounds. We prove online concentration bounds for the Clipped IS-based estimator (Theorems \ref{lem:clipped},\ref{thm: estimator concentrations}) that are valid under any \textit{arbitrary causal policy}, that is any policy that chooses experts based purely on observations made in the past. We achieve this by analyzing a carefully constructed martingale and show that the mean of the estimator concentrates around the true mean of the expert exponentially fast, similar to the deterministic sample setting above due to the information leakage across experts . To the best of our knowledge, these types of results have not been established and our technique may be of independent interest.

\noindent{\bf 3. Episodic behavior with bootstrapping:} In Section \ref{sec: episodic setting}, we also specify the construction of the approximate experts used by ED-UCB in the case when the supports of $X,V$ are finite in the episodic setting. We show that if the agent is bootstrapped with $\mathcal{O}(|\calv|\log(T) + \log(|\calx|NT\sqrt{E}))$ samples per expert, the use of ED-UCB over $E$ episodes, each of length $T$ guarantees a regret bound of $\mathcal{O}\left( E(N+1) + \nicefrac{N\sqrt{E}}{T^2}\right)$ where the dominant term does not scale with $T$. These are presented in Theorem \ref{thm: regret with bootstrapping} and Corollary \ref{thm: full regret online sampling} respectively. Our regret bound lies in between those of D-UCB in the full information setting and naive optimistic bandit policies (e.g., UCB in \cite{auer2002finite}, KL-UCB in \cite{garivier2011kl}), demonstrating the merits of sharing information among experts. We also mention how our methods can be extended to continuous context spaces in Section \ref{sec: discussions}. 

\noindent{\bf 4. Empirical evaluation:} We validate our findings empirically through simulations on the Movielens 1M \cite{harper2015movielens}and CIFAR10 \cite{krizhevsky2009learning} datasets in Section \ref{sec: experiments}. We provide soft-control on the genre of the recommendation for users that are clustered according to age and show that the performance of ED-UCB is comparable to that of D-UCB and significantly better than the naive strategies for multi-armed bandits.

\subsection{Related work}
Adapting to changing environments forms the basis of meta-learning \cite{thrun1998lifelong, baxter1998theoretical} where agents learn to perform well over new tasks that appear in phases but share underlying similarities with the tasks seen in the past. Our approach can be viewed as an instance of meta-learning for bandits, where we are presented with varying environments in each episode with similarities across episodes. Here, the objective is to act to achieve the maximum possible reward through bandit feedback, while also using the past observations (including offline data if present). This setting is studied in \cite{azar2013sequential} where a finite hypothesis space maps actions to rewards with each phase having its own true hypothesis. The authors propose an UCB based algorithm that learns the hypothesis space across phases, while quickly learning the true hypothesis in each phase with the current knowledge. Similarly, linear bandits where instances have common unknown but sparse support is studied in \cite{yang2020provable}. In \cite{cella2020meta,kveton2021meta}, meta-learning is viewed from a Bayesian perspective where in each phase an instance is drawn from a common meta-prior which is unknown. In particular, \cite{cella2020meta} studies meta-linear bandits and provide regret guarantees for a regularized ridge regression, whereas \cite{kveton2021meta} uses Thompson sampling for general problems, with Bayesian regret bounds for K-armed bandits. 

Collective learning in a {\bf fixed} and contextual environment with bandit feedback, where the reward of various arms and context pairs share a latent structure is known as Contextual Bandits (\cite{auer2002nonstochastic,chu2011contextual, bubeck2012regret,langford2007epoch, dudik2011efficient,agarwal2014taming,simchi2020bypassing,deshmukh2017multi} among several others), where actions are taken with respect to a context that is revealed in each round. In various works \cite{agarwal2014taming,simchi2020bypassing,foster2020beyond,foster2018practical}, a space of hypotheses is assumed to capture the mapping of arms and context pairs to reward, either exactly (realizable setting) or approximately (non-realizable), and bandit feedback is used to find the true hypothesis which provides the greedy optimal action, while adding enough exploration to aid learning. 

In the context of online learning, Importance Sampling (IS) is used to transfer knowledge about random quantities under a known target distribution using samples from a known behavior distribution in the context of off-policy evaluation in reinforcement learning \cite{mahmood2014weighted}. Clipped IS estimates are also commonly studied in order to reduce the variance of the estimates by introducing a controlled amount of bias \cite{charles2013counterfactual, bubeck2013bandits, lattimore2016causal, sen2017identifying}. Bootstrapping from prior data has been used in \cite{zhang2019warm, foster2018practical} to warm-start the online learning process.

Meta-learning algorithms take a model-based approach, where the invariant-structure (hypothesis space in \cite{azar2013sequential} or meta-prior in \cite{cella2020meta,kveton2021meta}) is first learnt to make the optimal decisions, while most contextual bandit algorithms are policy-based, trying to learn the optimal mapping by imposing structure on the policy space. Our approach falls in the latter category of optimizing over policies (aka experts) from a given finite set of policies. However, contrary to the commonly assumed deterministic policies, each policy in our setting is given by fixed distributions over arms conditioned on the context (learnt by bootstrapping from offline data). Using the estimated experts, in each episode (where both the arm reward per context and context distributions change), we quickly learn the average rewards of the experts by collectively using samples from all the experts. 

Previously in \cite{sen2018contextual}, we studied the non-episodic version of this problem in the full-information setting where expert policies and context distributions are known to the agent. The strategy presented therein guarantees a worst-case regret that scales logarithmically in the time-horizon. Their analysis built upon the results of \cite{sen2017identifying} that studied a best-arm identification variant of the problem.  In this work, we tighten the existing guarantee and prove a constant (time-independent) regret bound. Further, we provide techniques that reduce the computation complexity to be logarithmic in the number of experts per time (previously linear in the number of experts per time). We then generalize the full information setting to the case of empirical experts and unknown context distributions and also study the episodic version of the problem. 

\section{Problem Setting}
\label{sec:defs}

An agent interacts with a contextual bandit environment through a set of experts $\Pi = \{ \pi_1,... \pi_N\}$. At each time $t$, the agent observes a context $X_t \in \calx$ drawn independently from a fixed but unknown distribution $p(\cdot)$. The agent then selects an expert $\pi_{k_t}$ (or simply, expert $k_t$) that recommends an action $V_t$ from a finite set of actions $\calv$. The agent then receives a reward $Y_t\in[0,1]$ distributed according to $q(\cdot|X_t,V_t)$.

Given a context $X$, the action recommended by expert $i$ is sampled from the conditional distribution $\pi_i(\cdot|X)$. The choice of expert at this time $t$ can be instructed by the set of historical observations $(X_n,k_n, V_n,Y_n)_{n<t}$ and the current context $X_t$. We assume that the agent is only given access to \textit{all} conditional distributions in $\Pi$, while the context and reward distributions are \textit{unknown}.

\noindent{\bf Regret: }The objective of the agent in our contextual bandit problem is to perform competitively against the `best' expert in $\Pi$. We define $p_k(x,v,y) \triangleq p(y \vert v,x ) \pi_k(v \vert x)p(x)$ as the distribution of the corresponding random variables when the expert chosen is $\pi_k \in \Pi$. The expected reward of expert $k$ is then denoted by,
$\mu_k = \EE_{k}[Y],$ where $\EE_{k}$ denotes expectation with respect to distribution $p_k(\cdot)$. The best expert is given by $k^* = \argmax_{k \in [N]} \mu_k$. 

The goal of the agent is to minimize the \textit{regret} till time $T$, which is defined as $R(T) = \sum_{t = 1}^{T} \mu^* -\EE\left[\mu_{k_t} \right]$, where $\mu^* = \mu_{k^*}$. Note that this is analogous to the regret definition for the \textit{deterministic expert} setting of \cite{langford2008epoch}. We use $\Delta_{k} \triangleq \mu^* - \mu_k$ as the optimality gap in terms of expected reward, for expert $k$. Let $\pmb{\mu} \triangleq \{\mu_1,...,\mu_N \}$. We further assume that for all $i \in [N]$, $\mu_i \geq \gamma$. 

\noindent\textbf{Remark:} In contextual bandits, the best arm can change with the revealed context. However, learning this context-dependent best arm is often not tractable in practice when the arm/context spaces are large. As an alternative, access to a set of functions (or experts), each mapping contexts to actions, is assumed and the learner is now tasked with competing with the best expert in this given set \cite{auer2002nonstochastic, agarwal2014taming, syrgkanis2016efficient, simchi2020bypassing}. This notion of regret is especially useful when the number of experts $N$ is much smaller than the $|\calv|^{|\calx|}$, where $\calv$ is the set of arms and $\calx$ is the set of contexts. Such experts are usually learned using offline data and in combination with domain specific knowledge. For more details around this, we refer the reader to Chapter 18 of \cite{lattimore2020bandit}.

\section{Clipped Importance Sampling-based Estimator}\label{sec:DUCB}

The experts being modeled as conditional distributions over arms given contexts allows us to leverage Importance Sampling (IS) in order to use rewards collected under one expert to estimate the rewards of all other experts. Mathematically, the expected reward of expert $k$ can be written as

\begin{align}
	\label{eq:infoleakage}
	\mu_k = \EE_k[Y] = \EE_{j} \left[ Y \frac{\pi_k(V \vert X)}{\pi_j(V \vert X)}\right]. 
\end{align}

Note here that after the second equality, the expectation is taken under expert $j$ and the rewards are re-weighted appropriately. This has been termed as \textit{information leakage} and has been leveraged before in the literature of best-arm identification~\cite{sen2017identifying, lattimore2016causal,bottou2013counterfactual}.

The equation above is only valid for infinitely many samples from expert $j$. Further, the small values of the denominator $\pi_j(V|X)$ can introduce large variances in a standard empirical estimator of Equation \ref{eq:infoleakage}. To workaround these issues, we use a Clipped IS based estimator, similar to that introduced in \cite{sen2017identifying}.

We first define an $f$-divergence metric that is crucial to the design and analysis of this estimator. We define the conditional $f$-divergence as follows: 

\begin{definition}
	Let $f(\cdot)$ be a non-negative convex function such that $f(1) = 0$.
	For two joint distributions $p_{X,Y}(x,y)$ and $q_{X,Y}(x,y)$ (and the
	associated conditionals), the conditional $f$-divergence $D_{f}(p_{X
		\vert Y} \Vert q_{X \vert Y})$ is given by: $D_{f}(p_{X \vert Y} \Vert q_{X \vert Y}) = \EE_{q_{X,Y}} \left[f \left( \frac{p_{X \vert Y}(X \vert Y)}{q_{X \vert Y}(X \vert Y)} \right)\right].$
\end{definition}

Recall that $\pi_i$ is a conditional distribution of $V$ given $X$. Thus, $D_f(\pi_i \Vert
\pi_j)$ is the conditional $f$-divergence between the conditional distributions $\pi_i$ and $\pi_j.$ We now introduce the $M_{ij}$ measure:

\begin{definition}
	\label{def:mij}
	($M_{ij}$ measure)~\cite{sen2017identifying} Consider the function $f_1(x) = x \exp (x-1) -
	1$. We define the following log-divergence measure: $M_{ij} = 1 +
	\log (1 + D_{f_1} (\pi_i \lVert \pi_j)),$ $\forall i,j
	\in [N].$ 
\end{definition}

We also assume that the divergence between any two experts is upper bounded by $M<\infty$, i.e., $M\geq \max_{i,j\in[N]} M_{ij}.$ This immediately implies that every expert in $[N]$ recommends \textit{every} action in $\calv$ with a non-zero probability for \textit{every} context in $\calx$.

In order to define our Clipped IS-based estimator, we recall that the history of observations until time $t$ includes the set $\{X_n,k_n, V_n, Y_n\}_{n<t}$ and the context $X_t$. To ease notation, we will write $r_{i{k_t}}(t) = \frac{\pi_i(V_t|X_t)}{\pi_{k_t}(V_t|X_t)}$ to be the IS ratio between expert $i$ and the expert $k_t$ chosen at time $t$.

The estimator for the mean reward of expert $i$ at time $t$ is defined as:

\begin{align*}\label{eq:est1}
	\hat{\mu}_{i}(t) &= \frac{1}{Z_{i}(t)} \sum_{s=1}^t \frac{Y_s}{M_{ik_s}}r_{ik_s}(s)
	\cdot \ind\left\{  r_{ik_s}(s) \leq 2\log\left(\frac{2}{\epsilon(t)}\right)M_{ik_s}\right\}. \numberthis
\end{align*}

Here, $Z_i(t) = \sum_{j} N_j(t)/M_{ij}$ is the normalizing constant for expert $i$ with $N_j(t)$ as the number of times expert $j$ has been selected by time $t$. The bias-variance trade-off is controlled by the adjustable term $\epsilon(t)$. Formally, we define $\epsilon(t) = C w\left(\frac{\sqrt{t\log t}}{Z_i(t)}\right)$ where C is a constant. The function $w(\cdot)$ is defined as $w(x) = y \iff y/\log(2/y) = x$.

\noindent{\bf Intuition:} The clipped IS estimator is a weighted average of the samples collected under different experts, where each sample is scaled by the importance ratio as suggested by~\ref{eq:infoleakage}. At each time $t$, the adjustable term $\epsilon(t)$ is calculated to be used in the clipper levels. Then, the estimator is recomputed to include the re-weighted samples collected under other experts in the past which fall below the new clipper levels. Observe here that since $\epsilon(t)$ decreases with time, the clipper levels are increasing and thus, this estimator is asymptotically consistent. 

Compared to the vanilla IS estimator in Equation \ref{eq:infoleakage}, the clipped IS estimator above drops samples with large importance sampling ratios (due to the indicator function), leading to a biased estimate of the true mean $\mu_i$. However, as observed by authors in \cite{bottou2013counterfactual}, adding a controlled amount of bias to an importance sampling estimator helps its concentration behavior due to reduced variance. This variance reduction is thanks to the fact that the clipping leads to an estimate bounded within a smaller range compared to the vanilla (potentially unbounded) estimator. Additionally, the clipper level values and the weights are dependent on the divergence terms $M_{ij}$'s. When the divergence $M_{ij}$ is large, it means that the samples from expert $j$ is not valuable for estimating the mean for expert $i$. We will show in Theorem \ref{lem:clipped} below that this leads to exponential concentrations of the clipped IS estimate $\hat\mu_i(t)$ around the true mean $\mu_i$.

\section{Divergence-based Upper Confidence Bound Algorithm}

Recall that the goal of the agent is to remain competitive with the best expert in the given set of experts $\Pi$. We propose an optimistic index-based algorithm motivated by the popular Upper Confidence Bound (UCB) algorithm for stochastic Multi-armed Bandits \cite{auer2002finite}. Our algorithm uses the Clipped IS estimators developed in the previous section in order to compute a high-probability upper confidence bound for the mean of each expert. At each time, the policy chooses experts greedily according to these UCB's. The Divergence-based UCB algorithm (or simply, D-UCB) is summarized in Algorithm \ref{alg:DUCB}.

\begin{algorithm}
	\begin{algorithmic}[1]
		\STATE For time step $t = 1$, observe context $X_1$ and choose a random expert $\pi \in \Pi$. Play an arm drawn from the conditional distribution $\pi(V \vert X_1)$. 
		\FOR {$t = 2,...,T$}
		\STATE Observe context $X_t$
		\STATE Let $k_t = \argmax_{k} U_{k}(t-1) \triangleq \hat{\mu}_k(t-1) + s_k(t-1)$. 
		\STATE {Sample action $V_t$ from the distribution $ \pi_{k_t} (\cdot \vert X_t)$.} 
		\STATE Observe the reward $Y_t$.
		\ENDFOR
	\end{algorithmic}
	\caption{D-UCB: Divergence based UCB for contextual bandits with stochastic experts}
	\label{alg:DUCB}
\end{algorithm}

Here, the confidence bonus in Algorithm~\ref{alg:DUCB} for the estimator $\hat{\mu}_{k}(t)$ at time $t$ is chosen as $s_{k}(t) = \frac{3}{2}\epsilon(t).$

\subsection{Regret of D-UCB}\label{sec: full info regret}

In this section, we discuss the performance of D-UCB. In order to upper bound the expected regret incurred by our algorithm by time $T$, we require concentration guarantees for the estimators $\hat\mu_{k}(T)$. In \cite{sen2017identifying}, the authors provide concentration bounds for these estimators assuming that the number of times an expert is played by time $T$ is known. However, in our case, since the experts are chosen in an online fashion, the number of times an expert is chosen by time $T$ is a random variable upper bounded by $T$. Since the existing analysis does not apply in this case, we prove the following  Lemma using martingale concentrations in place of the Chernoff bounds used previously to account for the random nature of expert plays. This lemma provides exponentially fast concentrations for the estimator in Equation \ref{eq:est1} around the true mean of the corresponding expert.

\begin{theorem}\label{lem:clipped}
	For any expert $j\in [N]$, the estimator $\hat\mu_j(t)$ defined in Equation \ref{eq:est1} satisfies
	\begin{align*}
		&\PP\left( (1 - \beta(t))\left(\mu_j - \frac{\epsilon(t)}{2}\right) \leq \hat{\mu}_j(t) \leq (1 + \beta(t)) \mu_j\right ) \geq 1 - 2\exp \left( - \frac{\gamma^2 \beta(t)^2t}{128 M^2 (\log (2 / \epsilon(t)))^2 }\right).
	\end{align*}
	when $\beta(t)$ and $\epsilon(t) < \gamma$ are fixed non-negative constants and $M\geq \max_{i,j} M_{ij}$ is the \textit{finite} upper bound on the divergence between any two experts. 
\end{theorem}

By design, in the D-UCB (Algorithm \ref{alg:DUCB}), the mean estimates of \textit{all} experts are updated at each time. This departs from the stochastic bandit variant where the mean of \textit{only} the played arm is updated. Therefore, we expect that after sufficient time has passed, the mean estimates of all the experts are close to their true counterparts. Indeed, we show that this intuition holds and to this end, we define a series of problem-dependent times $\tau_i$ for all $i\in[N]$ as follows:
\begin{align}\label{eq:Tk defs}
	\tau_1 = \min\left\{ t: Cw\left( \sqrt{\log t/t} \right) \leq \gamma  \right\},~~~~
	\tau_k = \min\left\{ t\geq \tau_1: \frac{t}{\log t} \geq \frac{9C^2M^2 \log^2 \left( \nicefrac{6C}{\Delta_k}\right) }{\Delta_k^2} \right\}.
\end{align}

\begin{remark}
	We note here that the times $\tau_i$ for all $i\in[N]$ defined above are {\bf\em deterministic constants} that do not scale with time $t$. In particular, these are not the random number of times expert $i$ is played, which is the notation commonly used in bandit literature.
\end{remark}

Recall here that $\Delta_k = \mu^* - \mu_k$ is the suboptimality gap of expert $k$, $M= \max_{i,j\in[N]}M_{ij}$. Without loss of generality, we assume that the experts are indexed such that $0 = \Delta_1 < \Delta_2 \leq \Delta_3, ... \Delta_N$ and thus, $\tau_N\leq \tau_{N-1}...\leq \tau_2$.

With $\tau_k$ as defined above, using the results of Theorem \ref{lem:clipped}, we can establish the following statements:
\begin{enumerate}
	\item[1.] For all $t\geq \tau_1$, $\prob(U_{k^*}(t) \leq \mu^*)\leq t^{-2}.$
	\item[2.] For any $k:\Delta_k>0,$ for all $t\geq \tau_k$, $\prob(U_k(t)\geq \mu^*)\leq t^{-2}.$
\end{enumerate}
Together, these two statements give us the following corollary.
\begin{corollary}\label{cor: prob of arm play}
	For any suboptimal expert $k:\Delta_k>0$, at any time $t\geq \tau_k$, $\prob(k_t = k) \leq \nicefrac{2}{t^2}$.
\end{corollary}

Since the above is true for each time $t$ after $\tau_k$, since $\lim_{T\rightarrow\infty}\sum_{t=1}^T \nicefrac{1}{t^2} = \nicefrac{\pi^2}{6}$, we can show that expert $k$ is only chosen a constant number of times from then on. Extending this to argument to all suboptimal experts, we can conclude that after time $\tau_2$, the optimal expert is played all but a constant number of times. This leads to our main \textit{constant regret} result below:
\begin{theorem}\label{thm: one episode regret}
	The regret incurred by Algorithm \ref{alg:DUCB} by time $T$ is upper bounded by
	\begin{align*}
		R(T) &\leq \frac{\pi^2}{3} \sum_{k=2}^N \Delta_k + \tau_N\Delta_N + \sum_{k=2}^{N-1} \left( \tau_k - \tau_{k+1}\right) \Delta_k.
	\end{align*}
\end{theorem}

\noindent\textbf{Remarks:} \textbf{1. Value of $C$:} In the above, we use $C = \nicefrac{16M}{\gamma}$ with $\gamma$ the lower bound on the reward of all experts and $M$ the upper bound on the divergence between any pair of experts. However, our empirical evaluations show that we observer that smaller values of $C$ also lead to constant regret.\\
\noindent\textbf{2. About $M$:} We have assumed that all experts recommend all actions with non-zero probability, guaranteed by $M<\infty$. While this assumption leads to clean analysis for constant regret, it is not clear if it is necessary. For example, suppose there exist an action $v\in\calv$ that is only ever played by only one expert in $[N]$ with a low probability under any context $x\in \calx$. It is reasonable that the effect of this action on the mean reward of this expert is low and can be upper bounded. Hence, we might be able to recover constant regret by simply ignoring this action in the cases where the rewards obtained by this action are low. This would require a more careful regret analysis that builds on the methods used to prove Theorem \ref{thm: one episode regret}.\\
\noindent\textbf{3. Comparing to Linear Bandits:} The structure we impose on the experts and their interactions with the actions leads to the leakage of information across experts. Indeed, other settings, most famously that of linear bandits \cite{li2010contextual}, also share similarities in that playing one action leads to non-trivial information about other actions. However, linear bandit-like formulations can not achieve the constant regret guarantees we are able to provide. This is due to the fact that in our setting, playing one arm explicitly generates a `pseudo-sample' for \textit{every} other arm using the clipped importance sampling estimators. When the clipper levels are large enough, this leads to us essentially operating in the full-information setting, albeit with a potentially larger variance of reward per expert. However, in a linear bandit, this property can not be guaranteed --- firstly, there can be no information leakage in orthogonal directions and further, rewards from one arm can not be scaled uniformly to infer rewards from another.

\subsection{Reducing Computation}\label{sec: improved scaling}

In each round $t$ of the D-UCB algorithm (Algorithm \ref{alg:DUCB}, the agent computes mean estimates of each of the $N$ experts using all the observations up to time $t-1$ in order to form the respective UCB indices. Recall that the clipper levels of the estimator defined in Equation \ref{eq: empirical IS estimator} are increasing over rounds. Hence, updating these at each round is computationally expensive. To workaround this cumbersome update procedure, we introduce a variant of D-UCB called D-UCB-lite in Algorithm \ref{alg:DUCB-lite}. Here, at each time, the algorithm updates only a small subset of the experts using only a fraction of the collected samples. The rationale being that if we carefully choose the rate at which each arm is updated over time, then provided enough samples, the algorithm should still be able to distinguish between the true best expert and the remainder. 

\begin{algorithm}
	\begin{algorithmic}[1]
		\STATE For time step $t = 1$, observe context $X_1$ and choose a random expert $\pi \in \Pi$. Play an arm drawn from the conditional distribution $\pi(V \vert X_1)$.
		\STATE For each expert, set $B_k(1) = U_k(1)$ and initialize $N_k(t) = 1$.
		\FOR {$t = 2,...,T$}
		\STATE Observe context $X_t$
		\STATE Let $k_t = \argmax_{k} B_{k}(t-1)$.
		\STATE {Sample action $V_t$ from the distribution $ \pi_{k_t} (\cdot \vert X_t)$, observe the reward $Y_t$.}
		\STATE Sample a subset $S'(t)$ of experts of size $\log N -1$ uniformly from $[N]\backslash \{k_t\}$.
		\FOR {$k \in S'(t)\cup k_t$}
		\IF {$N_k(t)\leq \lfloor t^{\frac{1}{\beta}}\rfloor$}
		\STATE $B_k(t) = U_k(\lfloor t^{\frac{1}{\beta}}\rfloor)$, $N_k(t) = \lfloor t^{\frac{1}{\beta}}\rfloor$.
		\ELSE{} 
		\STATE $~B_k(t) = B_k(t-1)$.
		\ENDIF
		\ENDFOR
		\ENDFOR
	\end{algorithmic}
	\caption{D-UCB-lite: D-UCB with intermittent updates}
	\label{alg:DUCB-lite}
\end{algorithm}

This algorithm is similar to D-UCB in all aspects other than the index update rule. In contrast, D-UCB-lite first samples a $(\log N -1)$-sized subset $S'(t)$ of experts at time $t$. For each expert, it maintains a counter $N_k(t)$ which is updated to be the number of samples used to build a current index of expert $k$. The algorithm then updates the indices of the chosen expert $k_t$ and those in $S'(t)$ only if they have not used $\lfloor t^{\frac{1}{\beta}}\rfloor$ samples by time $t$ for some chosen $\beta\geq1$. In the worst case, this algorithm updates $\log N$ experts with $\lfloor t^{\frac{1}{\beta}}\rfloor$ samples each at every time step (rather than $N$ experts with $t$ samples each).

Specifically, at round $T$, D-UCB incurs a computational cost that scales as $\mathcal{O}(TN)$ This is because D-UCB needs to update the estimates of each of the $N$ experts at the new clipper level at time $T$ using all $T$ collected samples. In contrast, D-UCB-lite has a computational complexity of $\mathcal{O}(T^{\frac{1}{\beta}}\log N)$ at round $T$. The intermittent update strategy naturally loses performance compared to D-UCB. The following lemma specifies this loss.

\begin{lemma}\label{lem: improved computation}
	With $c = N/\log N$, define $\tau_{1,\beta} =\min\{t: t-2c\log t \geq \tau_1^\beta\}$ and for any $k\neq 1$, $\tau_{k,\beta} = \min\{t\geq \tau_{1,\beta}: t - 2c\log t \geq \tau_k^\beta\}$. The regret of D-UCB-lite in Algorithm \ref{alg:DUCB-lite} can be bounded as 
	\begin{align*}
		R(T) \leq \tau_{ N,\beta}\Delta_N + \frac{\pi^2}{3}\sum_{k=2}^N\Delta_k + \sum_{t = 1}^T \frac{2\Delta_N}{\lfloor t - 2c\log t \rfloor^{\frac{2}{\beta}}}.
	\end{align*}
\end{lemma}

Recall here that $\tau_i$ are defined in Equation \ref{eq:Tk defs}. The final summation converges for all values of $\beta<2$ recovering the time-independent regret guarantees as in Theorem \ref{thm: one episode regret}, albeit with a larger numerical value.

In Appendix \ref{app: improved scaling}, we also discuss a specific generative model under which our regret scales \textit{logarithmically} in the number of experts $N$ (as opposed to the linear scaling suggested by Theorem \ref{thm: one episode regret}).

\section{Empirical Experts and Unknown Context Distributions}\label{sec: empirical setting}

In the previous sections, we developed the clipped importance sampling estimator that relied crucially on the knowledge of the expert distributions $\pi_i(V|X)$ for all $i\in[N]$. Further, to compute the divergence metric $M_{ij}$, the agent also needs to use the context distribution chosen by nature. In practice, however, it is often the case that these two distributions are not readily accessible to the policy designer.

The most natural alternative to knowing the expert policies explicitly is to estimate them empirically using offline samples and use these estimates in D-UCB to compute the mean estimates. These expert policies appear in our mean estimators in Equation \ref{eq:est1} in the form of the IS ratios and are used to scale rewards as well as decide the state of the clippers. Therefore, to maintain our constant regret guarantee, we not only need to control for error in the estimate of $\pi_k(V|X)$, but also in the IS ratios $r_{ij}(V|X)$ for each pair of experts. 

Similarly, in the case where context distributions are unknown, empirical estimation can be carried out in order to produce estimates of the $M_{ij}$ divergence measures to be used in the reward estimators. Unfortunately, this is not feasible when the size of context set $\calx$ is large. To work around this, we will only assume access to a universal lower bound on the probability of occurrence of any context.

The rest of this section will be devoted to modifying the Clipped Importance Sampling estimator in Equation \ref{eq:est1} to use the empirical estimates of the corresponding expert policies.

\subsection{Empirical Clipped IS based estimator}\label{sec: empirical estimator}
We begin by defining our empirical expert policies and translating their error to the error in the IS ratios. Recall that we write $r_{ij}(V|X) = \nicefrac{\pi_i(V|V)}{\pi_j(V|X)}$ as the IS ratio between experts $i$ and $j$ and further abuse this by denoting $\hat{r}_{ij}(V|X)$ as the empirical IS ratio between experts $i$ and $j$. That is, for empirical estimates $\hat\pi_k$ for all $k\in[N]$, we have $\hat{r}_{ij}(V|X) = \nicefrac{\hat\pi_i(V|X)}{\hat\pi_j(V|X)}$.

\begin{proposition}\label{prop:ratio defs}
	Suppose $\xi$ to be the maximum error in the empirical expert policies, i.e., $\forall i\in[N], \|\pi_i - \hat{\pi}_i\| _\infty \leq \xi$. Additionally, let $\pi_i(V|X)\geq p_V >0$ for all $i,v,x$. Then, the following hold for all $(x,v)\in\calx\times\calv$:
	\begin{align*}
		r_{ij}(v|x) \geq \undr_{ij}(v|x) := \hat{r}_{ij}(v|x) - \frac{\xi}{p_V(p_V-\xi)},~~~~~~~~r_{ij}(v|x) \leq \overr_{ij}(v|x) := \hat{r}_{ij}(v|x) + \frac{\xi}{p_V(p_V+\xi)}.
	\end{align*}
\end{proposition}

The proof of the above follows from Lemma 5.1 in \cite{cayci2019learning}. We note here that the assumption of a universal lower bound $p_V$ on the probability mass $\pi_i(V|X)$ is necessary to guarantee finiteness of the IS ratio. If for any expert $j$ this does not hold, then given samples from this expert, we would not be able to infer anything meaningful about any of the other experts. This is also true for our discussions in the previous section when the expert policies were known completely where this assumption was implicit in that $M = \max_{i,j\in[N]}M_{ij}$ was assumed to be finite.

We now introduce the Empirical Clipped Importance Sampling based Estimator, similar to its non-empirical counterpart in Equation \ref{eq:est1}. To this end, we define the following natural lower bound to the $M_{ij}$ divergence measures in Definition \ref{def:mij}. 

\begin{align}
	\underline{D}_{f_1}(i||j) \triangleq p_X \sum_{x\in\calx}\sum_{v\in\calv} \left(\hat{\pi}_j(V|X) - \xi\right)f_1\left(\undr_{ij}(V|X)\right),~~~~\underline{M}_{ij} \triangleq 1 + \log(1 + \underline{D}_{f_1}(i||j)). \label{eq: divergence estimates}
\end{align} 

Recall here that $f_1(x) = x\exp(x-1)-1$. Additionally, $p_X > 0$ is the lower bound on the occurrence of any context, i.e., $p_X \geq \min_{x\in \calx} p(x)$. We are now ready to define our empirical reward estimator. Suppose $(X_s, k_s, V_s, Y_s)_{s<t}$ is the historical observations up to time $t$ and $X_t$ is the provided context. We define our empirical clipped IS estimator for the mean of expert $i$ at time $t$ as
\begin{align}
	\tilde{Y}_i(t) &= \frac{1}{Z_i(t)} \sum_{s=1}^t\left( \frac{Y_s}{\underline{M}_{ik_s}}\undr_{ik_s}(s) \cdot \ind\left\{ \overr_{ik_s}(s) \leq 2\log\left( \frac{2}{\epsilon_i(t)}\right)\underline{M}_{ik_s}\right\} \right). \label{eq: empirical IS estimator}
\end{align}

As before, we have $Z_i(t) = \sum_{s=1}^t   1/ \underline{M}_{ik_s}$ for $k_s$ the expert selected at time $s$ and $\epsilon(t)$ is the term that balances bias and variance (defined using this new version of $Z_i(t)$).

The empirical IS estimator $\tilde{Y}_k(t)$ is designed to be underestimate of the true mean $\mu_k$ on average. Additionally, its mean is also lesser than the mean of the full information IS estimator in Equation \ref{eq:est1}, making it further biased. We compensate for this by enlarging the confidence bonus ($s_k(t)$ in Section \ref{sec:DUCB}) by the maximum bias that $\tilde{Y}_k(t)$ can have by time $t$.

The UCB index of expert $i\in[N]$ at time $t$ is set to be 
\begin{align}
	U'_i(t) = \tilde{Y}_i(t) + e_i(t) + \frac{3}{2} \epsilon_i(t),~~~~ e_i(t) = \max_{\scriptscriptstyle j\in[N], (v,x)\in \calv\times\calx} \left[ \scriptstyle{\overr_{ij} - \undr_{ij}\ind \left\{ \overr_{ij} \leq 2\log\left( \frac{2}{\epsilon_i(t)}\right) \underline{M}_{ij}\right\}} \right].\label{eq: ucb index}
\end{align}

The index $U'_i(t)$ consists of three terms: the first is the estimate we define in Equation \ref{eq: empirical IS estimator}. The error term $e_i(t)$ captures the maximum error in our estimate due to the inherent inaccuracies in the empirical expert policies. Finally, the last term defines the length of the upper confidence interval which is characterized by $\epsilon_i(t)$ (which appears in the clipper level of the estimator). It can be shown that with high probability, this index $U'_i(t)$ is an over-estimate of the true mean $\mu_i$ of expert $i$. We summarize the Empirical D-UCB (ED-UCB) algorithm in Algorithm \ref{alg:EDUCB}.

\begin{algorithm}
	\begin{algorithmic}[1]
		\STATE {\bf Inputs:} Empirical experts $\hat\pi_i$ for $i\in[N]$ with maximum error $\xi$, lower bounds $p_X,p_V$.
		\STATE For time step $t = 1$, observe context $X_1$ and choose a random expert $\pi \in \Pi$. Play an arm drawn from the conditional distribution $\pi(V \vert X_1)$. 
		\FOR {$t = 2,...,T$}
		\STATE Observe context $X_t$
		\STATE Let $k_t = \argmax_{k} U'_{k}(t-1)$ with $U'_k(t)$ as in Equation \ref{eq: ucb index} 
		\STATE Observe the realization of the arm $V_t$ and reward $Y_t$.
		\ENDFOR
	\end{algorithmic}
	\caption{ED-UCB: Empirical D-UCB}
	\label{alg:EDUCB}
\end{algorithm}

\subsection{Regret Bounds}\label{sec: empirical regret}

The order of operations in proving regret bounds for ED-UCB are similar to that of D-UCB in \ref{sec: full info regret}. The key difference is that concentration bounds are now required on the estimator in Equation \ref{eq: empirical IS estimator} instead of its D-UCB counterpart that assumes access to all expert policies. We defer the development of these bounds to the appendix. As done previously, we then produce a sequence of times $\tau'_i$ for $i\in [N]$ as follows:
\begin{align*}\label{eq:Tk'defs}
	T_{clip} &= \min\{ t: \forall i,j\in[N], {\scriptstyle \overr_{ij} \leq 2\log\left( 2/\epsilon_i(t)\right)\underline{M}_{ij}}\}\\
	\tau'_1 &= \min\left\{ t \geq T_{clip}: {\scriptstyle Cw\left( \sqrt{\log t/t} \right) \leq \gamma } \right\},\\
	\forall i: \Delta_i>0, \tau'_i &= \min\left\{  t\geq \tau'_1: \scriptstyle{\frac{ t}{\log t} \geq \frac{9C^2M^2\log^2\left( \frac{6C}{(\Delta_i - \gamma p_V)} \right) }{\left(\Delta_i - \gamma p_V\right)^2} }\right\}.\numberthis
\end{align*}

These times can be used to a similar result to Corollary \ref{cor: prob of arm play}. Our final regret result follows:
\begin{theorem}\label{thm: per-epi regret main paper}
	Suppose the empirical estimation error $\xi$ is such that for all experts $i\in [N]$,
	\begin{align*}
		\| \pi_i - \hat\pi_i\|_\infty \leq \xi \leq 2\frac{\sqrt{1 + p_V^4\gamma^2} - 1}{p_V\gamma}.
	\end{align*}
	Consider $\tau'_k$ for $k\in [N]$ as defined in the above. Then, for $T\geq \tau'_2>0$, the expected cumulative regret of ED-UCB is bounded as
	\begin{align*}
		R(T) \leq \frac{\pi^2}{3} \sum_{k=2}^N \Delta_k + \tau'_N\Delta_N + \sum_{k=2}^{N-1} \left( \tau'_k - \tau'_{k+1}\right) \Delta_k := R(\mathbf{\Delta}),
	\end{align*}
	where $\mathbf{\Delta} = (\Delta_2, ... , \Delta_N)$ is the vector of suboptimality gaps.
\end{theorem}

\noindent{\bf Remarks: }\\
\noindent\textbf{1. Value of constants: } To analyze ED-UCB, we use $C = \nicefrac{32M}{\gamma(1-p_V)}.$ However, as with D-UCB, our empirical results suggest that smaller values suffice to achieve constant regret.

\noindent\textbf{2. Dependence on $N$ and intermittent updates:} We note here that all the discussions carried out in Section \ref{sec: improved scaling} also apply in the empirical case above. That is, ED-UCB also enjoys the improved scaling in $N$ under the generative model for suboptimality gaps and the computational complexity can be improved via the intermittent update strategy.

\noindent \textbf{3. Updating empirical expert policies online:} We assume that the empirical expert policies remain unchanged during the learning process. However, in practice, it is sensible to improve these estimates with the online collected data. Indeed, this can be done, but it makes analyzing the online regret more complex due to evolving expert policy estimates: The form of Theorem \ref{thm: one episode regret} depends on a universal error bound for all empirical expert estimates at a level $\xi$. By updating experts online based on collected samples, it is unclear how this error evolves with time as the rewards are random. We note that our regret result above is powerful as it implies that updating expert policy estimates online is \textit{not necessary to achieve constant regret}.

\noindent\textbf{4. Misspecifying $\xi$:} The legitimacy of the upper bound on expert policy estimation errors $\xi$ is imperative in order to guarantee the constant regret of ED-UCB. If this value is inaccurate, i.e., $\exists i\in[N]: \| \pi_i - \hat\pi_i\|_\infty >\xi$, then we can not guarantee that the UCB index of this expert is strictly below the true mean of the best expert (or above, if this was the best expert). Owing to this, we can no longer guarantee constant regret.

\section{Extending to Episodes}\label{sec: episodic setting}

In this section, we study the episodic setting. Here, the agent is to act on the environment for a total of $E$ episodes, each with $T$ time steps. In each episode, the distribution of contexts $p_e(x)$ as well as the reward distribution $q_e(y|v,x)$ is held fixed, but can change across episodes. Since the set of experts that the agent accesses remain fixed across episodes, the respective policies can be learnt offline and reused in each episode to guarantee constant regret via Theorem \ref{thm: per-epi regret main paper}. We recall our advertising example, where the agency learns the experts in its inventory before deployment and reduces the task of placing advertisements in each campaign to one of selecting the best expert from its roster. In most cases, learning these policies while the recommendations are generating rewards can be prohibitive due to the size of the context set $\calx$ and the action set $\calv$. However, in practice (including our advertising example), this learning can be carried out offline by repeatedly querying these experts for recommendations, thus decoupling this process with that of reward accumulation. 

Based on this observation, we suggest the use of basic sampling in order to build empirical expert policies that are bootstrapped into ED-UCB at the start of each episode. This choice helps us to implicitly handle the changes in the reward distributions in each episode. The reward distribution $q_e$ only affects the generated rewards $Y_s$ and the average rewards $\mu_{i,e}$, which affects the suboptimality vector $\mathbf{\Delta}_e$. The estimation procedure in Equation \ref{eq: empirical IS estimator} adapts to the former, while the latter simply causes the constant upper bound to vary with episodes. Thus, in what follows, we study the bootstrapping process and provide a corollary for the case when the expert policies are learnt online. We also abuse notation here by using subscripts of the episode index wherever necessary.

\subsection{Bootstrapping with offline sampling and online sampling}

In order to use empirical estimates in each episode, the agent must learn {\bf all} expert policies sufficiently well, characterized by the maximum error $\xi$. To this end, we assume that the agent is allowed to sample from a context distribution $p_{prior}(x)$ such that for any $x\in\calx$, $p_{prior}(x)\geq p_X >0$. In the context of our practical advertising example, this can also be thought of as trying to estimate the expert policies from a large pool of data collected from previous ad campaigns. Bootstrapping and warm-starting methods are common in contextual bandit algorithms; for examples, see \cite{zhang2019warm,foster2018practical} among others.

To achieve constant regret per-episode, Theorem \ref{thm: per-epi regret main paper} suggests the use of $\xi = 2\frac{\sqrt{1-p_V^4\gamma^2}-1}{p_V\gamma}$. Using Chernoff bounds, for a fixed expert, under a fixed context, collecting $n = \frac{2|\calv|\log(2T)}{\xi^2}$ would imply a maximum error of $\xi$ with probability at least $1-T^{-1}$ \cite{weissman2003inequalities}. However, since the sampling procedure is probabilistic through $p_{prior}(x)$, we have the following lemma:

\begin{lemma}\label{lem: samples lemma}
	Define $A = \frac{2n}{p_X} + \frac{\log(|\calx|NT\sqrt{E})}{2p_X^2}$. Let $\mathcal{E}$ be the event that after sampling each expert $i\in[N]$ $A$ times, we acquire at least $n$ samples under each context $x\in\calx$. Then, $\prob(\mathcal{E})\geq 1 - \frac{1}{T\sqrt{E}}$.
\end{lemma}

\begin{algorithm}
	\begin{algorithmic}[1]
		\STATE {\bf Inputs:} Sampling oracles for true agent policies $\pi_i(\cdot|\cdot)$, parameters $p_X, p_V, \gamma,E,T$.
		\STATE {\bf Bootstrapping:} Play each expert $A$ times to build approximate experts $\hat{\pi}_i(\cdot|x)$.
		\STATE {\bf Episodic Interaction:}
		\FOR {$e = 1,2,...E$}
		\STATE Play a fresh instance of ED-UCB (Algorithm \ref{alg:EDUCB}) with parameters $\hat{\pi}_i, p_V, p_X, \gamma$ for $T$ steps.
		\ENDFOR
	\end{algorithmic}
	\caption{Meta-Algorithm: ED-UCB for Episodic Bandits with Bootstrapping}
	\label{algo: ED-UCB for episodes}
\end{algorithm}

This lemma specifies that under the event $\mathcal{E}$, the maximum error in the empirical expert policies is bounded by $\xi$ with probability at least $1-T^{-1}$. The agent then instantiates ED-UCB at the start of each episode. This process is summarized in Algorithm \ref{algo: ED-UCB for episodes}. A straightforward application of the Law of Total Probability leads to the following theorem.

\begin{theorem}\label{thm: regret with bootstrapping}
	The regret of the agent in Algorithm \ref{algo: ED-UCB for episodes} is bounded as
	\begin{align*}
		R(T,E) \leq \sqrt{E} + E + \left(\sum_{e=1}^T R(\mathbf{\Delta_e})\right)\left(1 +\frac{1}{T^2\sqrt{E}} \right) = \mathcal{O}\left(E(N+1) + \frac{N\sqrt{E}}{T^2}\right).
	\end{align*}
\end{theorem}

Here, $R(\mathbf{\Delta}_e)$ is defined as the regret bound defined by Theorem \ref{thm: per-epi regret main paper} for episode $e$.

This result extends to the online setting naturally. In this case, the agent spends the first $AN$ time steps collecting samples and builds the empirical estimates of the experts. After this time, the agent continues as if it were bootstrapped. Since these expert policies do not change with episodes, the agent only incurs this additional $AN$ regret once. We summarize this in the following corollary.

\begin{corollary}\label{thm: full regret online sampling}
	The online estimation of the estimation oracles adds an additional regret of $AN$ to that in Theorem \ref{thm: regret with bootstrapping}. The total regret of the online process can be bounded as 
	\begin{align*}
		R(T,E) = {\mathcal{O}\left(N\log(NT^2\sqrt{E}) + E(N+1) + { \frac{N\sqrt{E}}{T^2}}\right)}.
	\end{align*}
\end{corollary}

\noindent\textbf{Remark:} As observed at the end of Section \ref{sec: empirical regret}, using online samples to improve estimates of expert policies in practice could lead to improved regret performance. Even if we were to improve these estimates at the end of each episode, the improvements would depend on the trajectory of observations that are random. Thus, for our analysis, we assume that empirical estimates are not updated after bootstrapping.

\section{Discussions}\label{sec: discussions}
\noindent{\bf How useful is information leakage: }The most natural alternative to considering the information leakage across the experts is to treat each of them independently. In this case, the problem reduces to a standard multi-armed bandit problem, where each expert is treated as an arm. In this case, the optimistic Upper Confidence Bound (UCB) strategies (as in \cite{auer2002finite, garivier2011kl}) explore each suboptimal expert $\log T$ times by time $T$, thus resulting in an overall regret that scales as $\mathcal{O}(N\log T).$ In contrast, leveraging the information leakage through Importance Sampling as in D-UCB does not necessitate any exploration beyond the time $\tau_1$(analogously $\tau'_1$ in the empirical formulation) and thus suffers regret that does not scale in the horizon. This carries on to the episodic setting, where naive UCB-type algorithms (that reset after each episode) incur a regret of the order $\mathcal{O}(EN\log T)$, but without the need for any bootstrapping.

\noindent\textbf{Lack of lower bound $p_V$:} Our assumption of the knowledge of $p_V$ allows us to develop IS estimates that are finite, leading to constant regret. If there exist a sub-optimal expert that does not satisfy this bound, it would have to be explored at a logarithmic rate since there is no information leakage with respect to this expert. However, samples could still be shared among the other experts through the use of D-UCB, making it a stronger baseline than naive policies. The question of optimality in this case remains open.

\noindent\textbf{Infinite context spaces:} The context distribution $p_e(x)$ has only been used the quantities $M_{ij,e}$ and the upper bound $M$ (Equation \ref{eq: divergence estimates}). For continuous contexts, the results in Section \ref{sec: empirical setting} can be extended by assuming knowledge of upper and lower bounds on the density function $p_e(x)$ and swapping out the summation for an integral in Equation \ref{eq: divergence estimates}. This can further be extended to the episodic case by assuming access to approximate oracles for expert policies.

\noindent\textbf{Stochastic episode lengths and unknown change-points:} Our analysis extends to the setting where the number of episodes and the length of each episode are random quantities with upper bounds $E,T$ respectively. In the case where the end of the episode is not communicated to the agent, ED-UCB can potentially be slow to adapt to the modified environment which leads to linear regret. Thus, tracking the best expert in unknown non-stationary environments is an important avenue for future work.

\noindent\textbf{Lower bounds:} The use of Importance Sampling in our estimators implies that samples collected under one expert also serve as samples (up to scaling and clipping) for all other experts. Intuitively, this implies that after some initial exploration, playing the best expert at each time should provide the agent enough evidence to discredit other experts. Therefore, in the single-episode setting, it is reasonable to expect a time-independent constant lower bound; the upper bounds of ED-UCB and D-UCB are in agreement of this intuition. We note however, that an algorithm that learns the best arm per-context can incur negative linear regret with respect to our defined best expert. Thus, producing lower bounds on regret in our setting and under stronger notions of the best expert, as well the number of bootstrapping samples necessary in the episodic case serve as two important avenues of future work.
\begin{figure}[t]
	\centering
	\includegraphics[height = 2.5in]{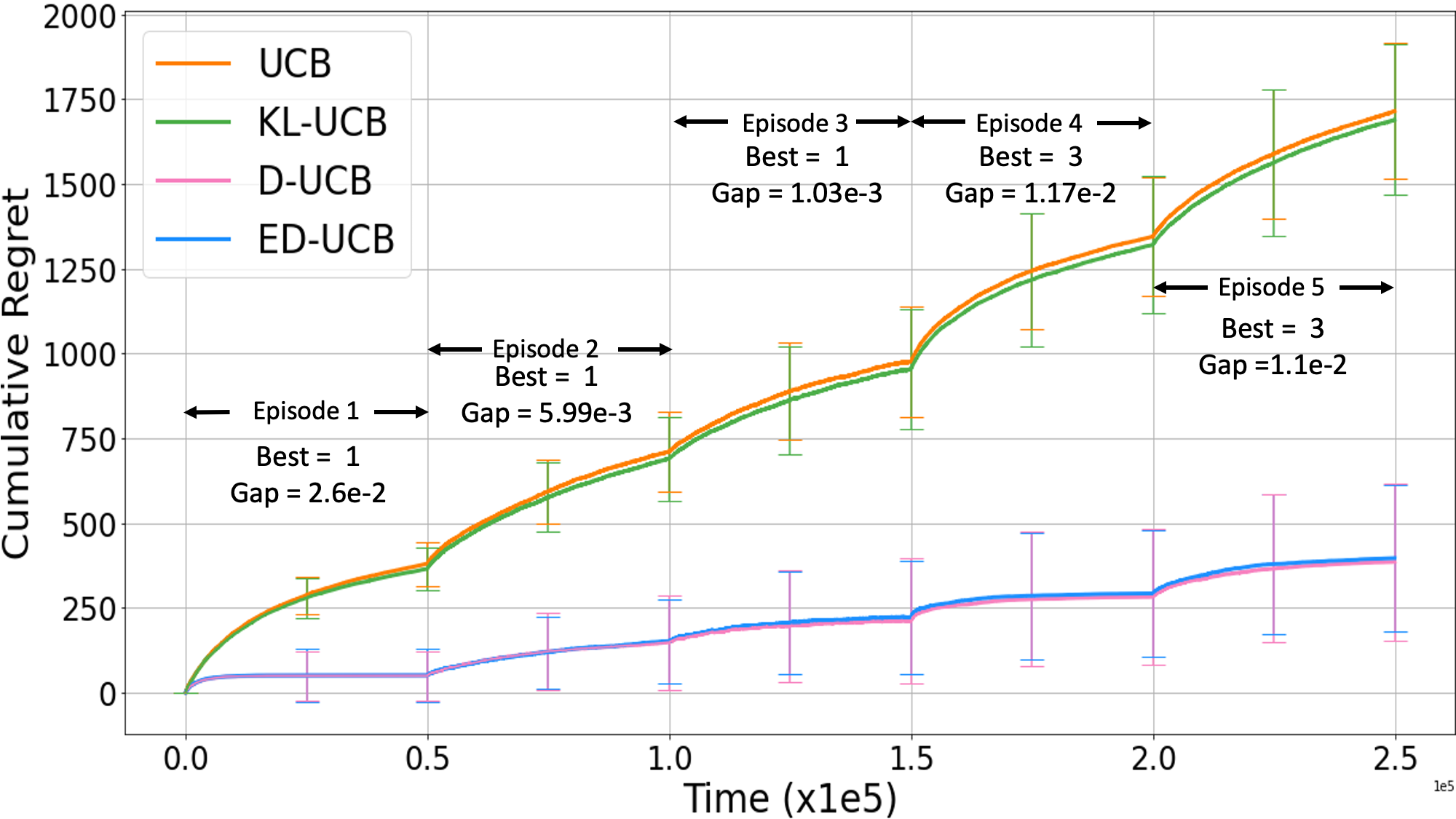}
	\caption{Experiments on the CIFAR-10 data set: The experiment consists of 5 episodes with $5\times10^5$ steps each. Plots are averaged over 300 independent runs, error bars indicate one standard deviation. Indices of the best expert and the minimum suboptimality gaps are presented.}
	\label{fig: cifar main figure}
\end{figure}

\begin{figure}[t]
	\centering
	\includegraphics[height = 2.5in]{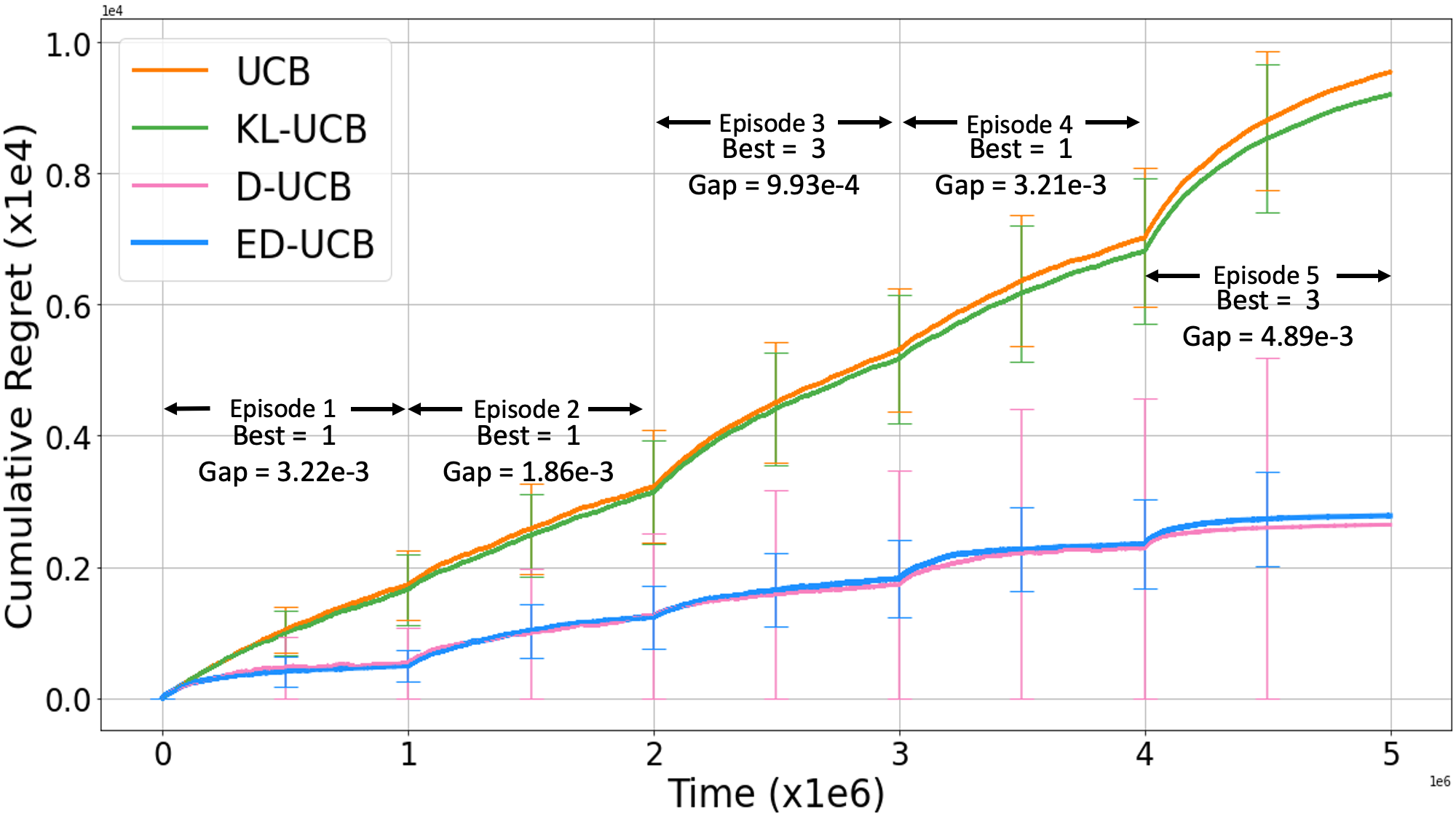}
	\caption{Experiments on the Movielens 1M data set: The experiment consists of 5 episodes with $10^6$ steps each. Plots are averaged over 100 independent runs, error bars indicate one standard deviation. Indices of the best expert and the minimum suboptimality gaps are presented.}
	\label{fig: movielens main figure}
\end{figure}
\section{Experiments}\label{sec: experiments}
We now present numerical experiments to validate our results above.  We build two sets of semi-synthetic experiments using the CIFAR-10 \cite{krizhevsky2009learning} and Movielens 1M \cite{harper2015movielens} data sets; we describe them below. We compare our D-UCB and ED-UCB algorithms with the naive UCB \cite{auer2002finite} and KL-UCB \cite{garivier2011kl} methods that do not leverage the information leakage. For all our experiments, we set the value of the constant $C=0.02$ for both D-UCB and ED-UCB. We use the best values of constants for the naive algorithms as suggested by \cite{lattimore2020bandit}.

\noindent\textbf{An image classification setup:} Using the CIFAR-10 dataset, we build 5 classifiers. Each of these is trained over data from 9 of the 10 labels, with a different label being omitted per classifier. We use these to build 5 experts: Given an input image, with probability 0.8, the expert recommends the class suggested by its associated classifier or recommends a uniformly random class with probability 0.2. This gives us experts that output each of the 10 classes with probability at least 0.02 for any image. Further, we use classes to form contexts as follows: for an image from class $c$, the context is provided as \texttt{``possibly an image of class $c$"}. We also sub-sample a set of 1000 test images (100 per class), called the `test set'.

Mapping this back to our episodic bandit setup, we have $|\calx| = 10$ contexts, $N=5$ experts and $|\calv|=10$ arms (or recommended classes). In the online phase, in each episode, we sample a context from the corresponding context distribution $X_t\sim p_e(x)$ and an image from the test set with this (possible) class uniformly at random. We choose an expert according to each of our evaluated algorithms and provide the recommended label as its output. The expert then observes a binary reward: 1 if the output was the true label of the image and 0 otherwise. We set $p_X = 0.05$ and use $p_V = 0.02$ to generate the necessary number of samples to form the empirical expert policies for ED-UCB. The results are averaged over 300 independent runs and presented in Figure \ref{fig: cifar main figure}

\noindent\textbf{A movie recommendation setup:}We use the Movielens 1M data set \cite{harper2015movielens} with 1 million ratings of approximately 3900 movies by 6000 users to construct a semi-synthetic bandit instance. First, we complete the reward matrix (scaled down to $(0,1)$) using the SoftImpute algorithm of \cite{mazumder2010spectral} included in the \texttt{fancyimpute} package \hbox{\cite{fancyimpute}}. We filter the number of movies to 618 using the completed matrix by eliminating ones that are mostly rated 0. Then, we cluster these movies based on 7 genres namely: \texttt{Action, Children, Comedy, Drama, Horror, Romance, Thriller} and users based on ages between $0-17, 18-24, 25-49, 49+$. At this stage, the average reward of all {\it (age,genre)} pairs are close to each other. To induce some diversity, we boost the rewards of the following {\it (age,genre)} pairs by 0.008: (0-17,\texttt{Children}), (18-24,\texttt{Horror}), (18-24,\texttt{Thriller}), (25-49,\texttt{Action}) and (25-49,\texttt{Drama}).

Experts are then randomly generated over the set of genres randomly with $p_V = 0.002$. Given an age group (context) and genre (expert-recommended action), a movie of the selected genre is picked uniformly and the reward is obtained from the completed reward matrix. We build empirical expert policies using Theorem \ref{thm: regret with bootstrapping} with $p_V = 0.2$ and a prior distribution satisfying $p_X = 0.05$ for $5$ episodes of length $10^6$ each. The averaged results of 100 independent runs are presented in Figure \ref{fig: movielens main figure}.

In both of Figures \ref{fig: cifar main figure} and \ref{fig: movielens main figure}, our Importance Sampling based policies show large improvements in regret over the naive baselines. In the movie recommendation case of Figure \ref{fig: movielens main figure}, the mismatch in $p_V$ in the sampling process causes the empirical estimates being formed with fewer samples than theoretically recommended. However, we empirically observe that ED-UCB still heavily outperforms the naive bandit policies and is comparable in regret to D-UCB. Some additional empirical results can be found in Appendix \ref{sec: additional exp}. These present some cases where the assumptions we make about the environment do not hold.

\section*{Acknowledgements}
We sincerely thank all the reviewers and editors for the feedback that helped improve the quality of the paper. This research was partially supported by NSF Grants 1826320, 2019844, 2107037 and 2112471, ARO grant W911NF-17-1-0359, US DOD grant H98230-18-D-0007, ONR Grant N00014-19-1-2566 and the Wireless Networking and Communications Group Industrial Affiliates Program.

\bibliographystyle{acm}
\bibliography{bibli}

\appendix

\section*{Appendix}
The appendix is structured as follows: We first discuss the proof of our regret results for ED-UCB(Algorithm \ref{alg:EDUCB}) in Section \ref{sec: empirical setting} at length. The results for D-UCB(Algorithm \ref{alg:DUCB}) in Section \ref{sec: full info regret} will then follow simply from these with similar arguments; we provide short proofs of these results. Next, we provide the proofs for our scaling and computational improvements discussed in Section \ref{sec: improved scaling}. Finally, we prove the episodic results from Section \ref{sec: episodic setting}.

\section{Useful Concentrations}
We begin by proving regret guarantees for ED-UCB in Algorithm \ref{alg:EDUCB} which will immediately imply all our results for D-UCB in Algorithm \ref{alg:DUCB}. We begin with some basic concentrations. First is a result about $L_1$ deviations of empirical probability distributions.

\begin{lemma}\label{lem: empirical probability concentrations}
	Let $p$ be a probability vector with $S$ points of support. Let $\hat{p} \sim \frac{1}{n} Multinomial(n,p)$ be an empirical estimate of $p$ using $n$ i.i.d. draws. Then, for any $S\geq 2$ and $\delta\in(0,1)$, it holds that
	\begin{align*}
		\prob\left({  \| p-\hat{p}\|_\infty \geq \sqrt{\frac{2S\log\left(\frac{2}{\delta} \right)}{n}}} \right) \leq \prob\left( { \| p-\hat{p}\|_1 \geq \sqrt{\frac{2S\log\left(\frac{2}{\delta} \right)}{n}}} \right) \leq \delta.
	\end{align*}
\end{lemma}
\begin{proof}
	The result follows from that of \cite{weissman2003inequalities} as $\|x\|_\infty \leq \|x\|_1$ for any vector $x$.
\end{proof}

Next is the proof of Proposition \ref{prop:ratio defs} which provides confidence bounds for ratios of random variables. We restate it here for convenience

\begin{proposition*}[Restatement of Proposition \ref{prop:ratio defs}]
	Suppose $\xi$ to be the maximum error in the empirical expert policies, i.e., $\forall i\in[N], \|\pi_i - \hat{\pi}_i\| _\infty \leq \xi$. Additionally, let $\pi_i(V|X)\geq p_V >0$ for all $i,v,x$. Then, the following hold for all $(x,v)\in\calx\times\calv$:
	\begin{align*}
		r_{ij}(v|x) \geq \undr_{ij}(v|x) := \hat{r}_{ij}(v|x) - \frac{\xi}{p_V(p_V-\xi)},~~~~r_{ij}(v|x) \leq \overr_{ij}(v|x) := \hat{r}_{ij}(v|x) + \frac{\xi}{p_V(p_V+\xi)}.
	\end{align*}
\end{proposition*}

\begin{proof}
	The proof follows the result from of Lemma 5.1 in \cite{cayci2019learning}. To ease notation, we fix an arbitrary $s\in \mathcal{S}$ and denote $\mu_k = \pi_k(s)$, $X_k = \hat{\pi}_k(s)$ for $k\in \{ i,j \}$. Under the event that $\|\mu_k - X_k\|_\infty\leq \xi$ for $k\in\{ i,j \}$, we have 
	\begin{align*}
		\frac{X_i}{X_j} &\geq {\frac{\mu_i -\xi}{\mu_j + \xi} = \frac{\mu_i}{\mu_j} - \frac{\xi}{\mu_j + \xi}\left( 1 + \frac{\mu_i}{\mu_j} \right)}\\
		&\geq{ \frac{\mu_i}{\mu_j} - \frac{\xi}{c + \xi}\left( 1 + \frac{1-c}{c}\right)}\\
		&= {\frac{\mu_i}{\mu_j} - \frac{\xi}{c(c+\xi)}}.
	\end{align*}
	
	The upper bound is proved similarly. Since the choice of $s\in \mathcal{S}$ was arbitrary and the event $\| \mu_k - X_k\|_\infty \leq \xi$ holds with probability at least $1-\delta$, the result follows.
\end{proof}

\section{Proofs of results in Section \ref{sec: empirical regret}}

\subsection{The simpler case of two experts}

We begin by considering 2 experts, $i,j$. We are given access to $t$ samples from expert $j$ and seek to estimate the mean of expert $i$ using the approximate policies $\hat{\pi}_i,\hat{\pi}_j$ with maximum error $\xi$. The arguments in this section closely follow the analysis of the two armed estimator in \cite{sen2017identifying}. We note here that we work with values of $\xi$ as in Theorem \ref{thm: per-epi regret main paper}; i.e., 
\begin{align*}
	{ \xi \leq 2 \frac{\sqrt{1+p_V^4\gamma^2}-1}{p_V\gamma} \iff \frac{\xi}{p_V}\left( \frac{1}{(p_V-\xi)} + \frac{1}{(p_V+\xi)}\right) \leq \frac{\gamma p_V}{2}}
\end{align*}
We refer the reader to Proposition \ref{prop:ratio defs} for the definitions of $\undr_{ij}(v|x), \overr_{ij}(v|x).$ Using the propsition, we also have
\begin{align*}
	\undr_{ij}(v|x) \leq \hat{r}_{ij}(v|x) \leq \overr_{ij}(v|x).
\end{align*}

For an arbitrarily chosen $\epsilon\in (0,1)$, we write
\begin{align}\label{eq:eta defn}
	\eta' &= \min\left\{a: \prob_i\left( \underline{r}_{ij}>a \right)\leq \frac{\epsilon}{2} \right\},\\
	\eta &= \min\left\{a: \prob_i\left( r_{ij}>a \right)\leq \frac{\epsilon}{2} \right\}.
\end{align}
One can easily check the following claim using basic properties of indicator functions:

\begin{claim}\label{claim: eta inequality}
	With $\eta, \eta'$ as above, for any $\epsilon\in(0,1),$ we have that $\eta'\leq\eta$. Further, $\ind\{r_{ij}\leq \eta \}\geq \ind \{ \overline{r}_{ij}\leq \eta' \}.$
\end{claim}

Our estimator for $\mu_i$ based on $t$ samples from expert $j$ is then defined as
\begin{align}\label{eq: 2-armed estimator version 1}
	\tilde{Y}_i(j,t) = \frac{1}{t} \sum_{s=1}^t Y_s \underline{r}_{ij}(s) \ind\left\{ \overline{r}_{ij}(s) \leq \eta' \right\}.
\end{align}

We recall the following result on the full information IS estimator (Lemma 1 in \cite{sen2017identifying}):

\begin{lemma}\label{lem: 2-armed previous bound}
	With $\eta$ as in Equation \ref{eq:eta defn} and the full information IS estimator as 
	\begin{align}\label{eq: previous 2-armed estimator}
		\hat{Y}_i^\eta(j,t) := \frac{1}{t} \sum_{s=1}^t Y_s r_{ij}(s) \ind \left\{ r_{ij}(s) \leq \eta \right\}
	\end{align}
	Then, for all $t\geq 1$, it holds that $\expec_j[\hat{Y}_i^\eta(j,t)] \leq \mu_i \leq \expec_j[\hat{Y}_i^\eta(j,t)] + \frac{\epsilon}{2}$.
\end{lemma}

We now compare our estimator in Equation \ref{eq: 2-armed estimator version 1} to the full information estimator in Equation \ref{eq: previous 2-armed estimator}. Due to Claim \ref{claim: eta inequality}, it holds trivially that $\tilde{Y}_i(j,t) \leq \hat{Y}_i(j,t)$. Consider the following chain:
\begin{align*}
	\hat{Y}_i(j,t) &= \hat{Y}_i(j,t) + \tilde{Y}_i(j,t) - \tilde{Y}_i(j,t)\\
	&\leq \tilde{Y}_i(j,t) + \frac{1}{t} \sum_{s=1}^t Y_s \left(\overline{r}_{ij}(s) - \underline{r}_{ij}(s)\ind\{\overline{r}_{ij}(s)\leq \eta'\} \right)\\
	&\leq \tilde{Y}_i(j,t) + \frac{1}{t} \sum_{s=1}^t \left(\overline{r}_{ij}(s) - \underline{r}_{ij}(s)\ind\{\overline{r}_{ij}(s)\leq \eta'\} \right)\\
	&\leq \tilde{Y}_i(j,t) + e_{ij}
\end{align*}

Here, the second inequality holds since $e_{ij}:= \max_{v,x}\overr_{ij} - \undr_{ij}\ind\{\overr_{ij}\leq \eta'\}\geq 0$ and $Y_s\leq 1$.
As a result of the arguments above, we have the following Lemma:
\begin{lemma}\label{lem: 2-armed interval version 1}
	With $\hat{Y}_i(j,t)$ and $\tilde{Y}_i(j,t)$ as defined in Equations \ref{eq: previous 2-armed estimator} and \ref{eq: 2-armed estimator version 1} respectively and $e_{ij}:= \max_{v,x}\overline{r}_{ij} - \underline{r}_{ij}\ind\{\overline{r}_{ij}\leq \eta'\}$, it holds that 
	\begin{align*}
		\expec_j[\tilde{Y}_i(j,t)]\leq \expec_j[\hat{Y}_i(j,t)]\leq \expec_j[\tilde{Y}_i(j,t)] + e_{ij}.
	\end{align*}
\end{lemma}
Together with Lemma \ref{lem: 2-armed previous bound} we have the following corollary:
\begin{corollary}
	The mean of the estimator $\tilde{Y}_i(j,t)$ according to the distribution of arm $j$ satisfies
	\begin{align}\label{eq: 2-armed estimator bound}
		\expec_j[\tilde{Y}_i(j,t)] \in \left[\mu_i - \frac{\epsilon}{2} - e_{ij} , \mu_i\right]
	\end{align}
\end{corollary}

\subsubsection{Simpler clipper levels}
In this section, we move from the abstract clipper levels in Equation \ref{eq:eta defn} to those based on divergences as in the estimator used in ED-UCB. To this end, using Markov's inequality, we have $\prob_i(\underline{r}_{ij}>a)\leq \exp(-a)\expec_i[\exp(\underline{r}_{ij})]$. Suppose that the RHS of the above is upper bounded by $\frac{\epsilon}{2}$. That is, $\exp(a) \geq \frac{2}{\epsilon}\expec_i[\underline(r_{ij})]$. Consider the following chain:
\begin{align*}
	\expec_i[\exp(\underline{r}_{ij})] = e + e\left(\expec_j[r_{ij}\exp(\underline{r}_{ij}-1) - 1]\right) \geq e + e\left(\expec_j[\underline{r}_{ij}\exp(\underline{r}_{ij}-1) - 1]\right) = e + e\underline{D}_{f_1}(\pi_i||\pi_j)
\end{align*}
Therefore, the RHS of the Markov inequality is upper bounded by $\frac{\epsilon}{2}$ if $\exp(a)\geq \frac{2}{\epsilon}(e+e\underline{D}_{f_1}(\pi_i||\pi_j))$ or, equivalently, if $a \geq \log\left( \frac{2}{\epsilon}\right)+\underline{M}_{ij}$. However, by definition, we must have that $\eta' \leq \log\left( \frac{2}{\epsilon}\right)+\underline{M}_{ij} \leq 2\log\left( \frac{2}{\epsilon}\right)\underline{M}_{ij}$.

Therefore, we redefine our original estimator in Equation \ref{eq: 2-armed estimator version 1} to use this new clipper level $\alpha_i(j,\epsilon):= 2\log\left( \frac{2}{\epsilon}\right)\underline{M}_{ij}$. We have
\begin{align}
	\tilde{Y}_i(j,t) := \frac{1}{t}\sum_{s=1}^t Y_s \underline{r}_{ij}(s) \ind\{ \overline{r}_{ij}(s)\leq \alpha_i(j,\epsilon)\}
\end{align}
We also restate Lemma \ref{lem: 2-armed interval version 1} for convenience:
\begin{lemma}\label{lem: final 2-armed lemma}
	For $\tilde{Y}_i(j,t)$ defined as above, and $e_{ij} = \max_{v,x} \overr_{ij}- \undr_{ij}\ind\{ \overr_{ij} \leq \alpha_i(j,\epsilon)\}$, 
	\begin{align}
		\expec_j[\tilde{Y}_i(j,t)] \in \left[ \mu_i - \frac{\epsilon}{2} - e_{ij}, \mu_i\right]
	\end{align}
\end{lemma}

\subsection{Estimator Concentrations}

The Clipped IS-based estimator with empirical policies is defined in Equation \ref{eq: empirical IS estimator}. To provide concentrations for this estimator, we define the following filtration by time $t$: $\mathcal{F}_t = \sigma\left(\{k_s, X_s, V_s, Y_s\}_{s=1}^{t-1}, k_t \right)$. Note that the filtration at time $t$ contains information about all the observations upto time $t-1$ as well as the choice of the arm at time $t$. We now define the following martingale that will be used to analyze the estimator:
\begin{align*}
	A_0 := 0, A_s := \sum_{l=1}^s \frac{L_i(l)}{\underline{M}_{ik_l}} - \sum_{l=1}^s \expec\left[\frac{L_i(l)}{\underline{M}_{ik_l}}\Big{|} \mathcal{F}_{l-1} \right]
\end{align*}
Where, to ease notation, we write $L_i(l) = Y_l \underline{r}_{ik_l}(l) \ind \{\overline{r}_{ik_l}(l)\leq \alpha_i(k_l,l) \}$. Since $\mathcal{F}_{l-1}$ contains knowledge of $k_l$, the second term in $A_s$ can be further simplified using $\mu_i(l) = \expec[L_i(l)|\mathcal{F}_{l-1}]$ as 
\begin{align*}
	A_s = \sum_{l=1}^s \frac{L_i(l)}{\underline{M}_{ik_l}} - \sum_{l=1}^s \frac{\mu_i(l)}{\underline{M}_{ik_l}}
\end{align*}

\begin{remark}\label{rem: intermediate bound for mu}
	Using Lemma \ref{lem: final 2-armed lemma}, we can write that $\mu_i(l) \in \left[ \mu_i - \frac{\epsilon}{2} - e_i(t),\mu_i\right] $ where $e_i(t) := \max_j e_{ij}$.
\end{remark} 
It is easy to see that $|A_s-A_{s-1}|\leq 4 \log\left(\frac{2}{\epsilon}\right)$. Therefore, using the Azuma-Hoeffding inequality for martingales with bounded differences, we can write:
\begin{align*}
	\prob\left( \Big|\sum_{l=1}^t \frac{L_i(l)}{\underline{M}_{ik_l}} - \sum_{l=1}^t \frac{\mu_i(l)}{\underline{M}_{ik_l}}\Big| \geq \beta \right) \leq 2\exp\left( -\frac{\beta^2}{32t\log^2\left(\frac{2}{\epsilon}\right)}\right)
\end{align*}

We are now ready to state and prove our concentration result.

\begin{theorem}\label{thm: estimator concentrations}
	The estimator $\tilde{Y}_i(t)$, as in Equation~\ref{eq: empirical IS estimator}, satisfies
	\begin{align*}
		\prob \left( \tilde{Y}_i(t) \notin \left[ (1-\beta) \left(\mu_i - \frac{\epsilon_i(t)}{2}- e_i(t)\right) , (1+ \beta) \mu_i\right] \right) \leq 2\exp\left( -\frac{t\beta^2\left(\gamma - 2e_i(t)\right)^2}{128M^2 \log^2\left(\frac{2}{\epsilon_i(t)}\right)}\right)
	\end{align*}
	for $t$ such that $\epsilon_i(t)\leq \gamma$, $\beta>0$ and $M = (1-p_X)|\calx||\calv|f_1\left(\nicefrac{p_V}{1-p_V}\right)$.
\end{theorem}

\begin{proof}
	\noindent\textbf{Upper Tail:} For any $\beta(t)>0$, we have the following chain:
	\begin{align*}
		(1+\beta(t))\left(\max_{l\in [t]} \mu_i(l)\right) \sum_{l=1}^t \frac{1}{\underline{M}_{ik_l}} &\geq \sum_{l=1}^t \frac{\mu_i(l)}{\underline{M}_{ik_l}} + \frac{t}{M}\times \beta(t) \times\max_{l\in[t]} \mu_i(l) \\
        &\geq \sum_{l=1}^t \frac{\mu_i(l)}{\underline{M}_{ik_l}} + \frac{\beta(t)t}{M}\left(\frac{\gamma}{2} - e_i(t)\right).
	\end{align*}
	The final inequality uses Remark \ref{rem: intermediate bound for mu} with $\epsilon(t) \leq \gamma\leq \mu_i$ for all $i\in \mathcal{A}$.
	Thus, we have that 
	
	\begin{align*}
		\prob\left(\tilde{Y}_i(t) \geq (1+\beta(t))(\max_{l\in [t]} \mu_i(l)\right) &\leq \prob\left( \sum_{l=1}^t \frac{L_i(l)}{\underline{M}_{ik_l}} \geq \sum_{l=1}^t \frac{\mu_i(l)}{\underline{M}_{ik_l}} + \frac{\beta(t)t}{M}\left(\frac{\gamma}{2} -e_i(t)\right) \right) \\
		&\leq \exp\left( -\frac{t\beta^2(t)\left(\frac{\gamma}{2} - e_i(t)\right)^2}{32M^2 \log^2\left(\frac{2}{\epsilon(t)}\right)}\right).
	\end{align*}
	
	\noindent\textbf{2. Lower Tail:}\\
	Again, for any $\beta(t)>0$,
	\begin{align*}
		(1-\beta(t))\left(\min_{l\in [t]}\mu_i(l)\right)\sum_{l=1}^t \frac{1}{\underline{M}_{ik_l}} &\leq \sum_{l=1}^t \frac{\mu_i(l)}{\underline{M}_{ik_l}} - \frac{\beta(t)t}{M}\left( \frac{\gamma}{2} - e_i(t)\right).
	\end{align*}
	Therefore, we have that 
	\begin{align*}
		\prob\left(\tilde{Y}_i(t) \leq (1-\beta(t))(\min_{l\in [t]} \mu_i(l))\right) &\leq \prob\left( \sum_{l=1}^t \frac{L_i(l)}{\underline{M}_{ik_l}} \leq \sum_{l=1}^t \frac{\mu_i(l)}{\underline{M}_{ik_l}} - \frac{\beta(t)t}{M}\left(\frac{\gamma}{2} -e_i(t)\right) \right)\\
		& \leq \exp\left( -\frac{t\beta^2(t)\left(\frac{\gamma}{2} - e_i(t)\right)^2}{32M^2 \log^2\left(\frac{2}{\epsilon(t)}\right)}\right).
	\end{align*}
	
	Combining the two tails above and using Remark \ref{rem: intermediate bound for mu}, we have
	\begin{align*}
		\prob\left(\tilde{Y}_i(t) \not\in \left[ (1-\beta(t)) \left({\scriptstyle \mu_i - \frac{\epsilon(t)}{2}- e_i(t)}\right) , (1+ \beta(t)) \mu_i\right]\right)\leq 2\exp\left( -\frac{t\beta^2(t)\left(\frac{\gamma}{2} - e_i(t)\right)^2}{32M^2 \log^2\left(\frac{2}{\epsilon(t)}\right)}\right).
	\end{align*}
\end{proof}

\subsection{Per-Expert Concentrations}

In this section, we provide concentration results for optimal and sub-optimal experts separately. We begin with the following claim:

\begin{claim}\label{claim: error bound}
	For all $t\geq T_{clip}$ and all $k\in [N]$, $e_k(t) = \frac{\gamma p_V}{2}$.
\end{claim}
\begin{proof}
	Fix $i,j\in [N]$ and $(v,x)\in \mathcal{V}\times\mathcal{X}$, arbitrary. Consider at some $t\geq T_{clip}$
	\begin{align*}
		&\overr_{ij}(v|x) - \undr_{ij}(v|x) \ind\left\{\overr_{ij}(v|x) \leq \alpha_{ij}(t)\right\}\\
		&~~~~= \overr_{ij}(v|x) - \undr_{ij}(v|x) ~~~~~(\text{By definition of $T_{clip}$})\\
		&~~~~= \frac{\xi}{p_V}\left( \frac{1}{(p_V-\xi)} + \frac{1}{(p_V+\xi)}\right) \leq \frac{\gamma p_V}{2} 
	\end{align*}
	Since this holds for the arbitrary choice of $i,j,v,x$, it must hold that $e_k(t)\leq \frac{\gamma p_V}{2}.$
\end{proof}

The following lemmas provide concentrations for the best expert and suboptimal experts:

\begin{lemma}\label{lem: best arm concentration}
	Let $\beta_{k^*}'(t) = Cw\left(\sqrt{\frac{\log t}{t}}\right)$ and $\tau'_1$ be as in Equation \ref{eq:Tk'defs}. Then, for all $t\geq \tau'_1$, it holds that $\prob(U'_{k^*}(t) \leq \mu^*) \leq \frac{1}{t^2}.$
\end{lemma}
\begin{proof}
	We have that for $t\geq \tau'_1,$
	\begin{align*}
		\prob({ U'_{k^*}\leq \mu^*}) &\leq \prob\left({\tilde{Y}_{k^*}(t) \leq \mu^* - \frac{3}{2}\beta'_{k^*}(t) - e_{k^*}}\right)\\
		&\leq \prob\left({ \tilde{Y}_{k^*}(t) \leq \mu^* - \mu^*\beta'_{k^*}(t) - (1-\beta'_{k^*}(t))\left(\frac{\beta'_{k^*}(t)}{2} + e_{k^*}\right)}\right)\\
		&=\prob\left(\tilde{Y}_{k^*}(t) \leq {(1-\beta'_{k^*}(t))\left(\mu^* -\frac{\beta'_{k^*}(t)}{2} - e_{k^*}\right)}\right).
	\end{align*}
	In the above, the first inequality follows since $\left( Z_k^*(t)\leq t, w(x) \text{ increasing}\implies \beta_{k^*}(t) \geq \beta'_{k^*}(t)\right)$ and the second is due to $\left( \beta'_{k^*}(t)\leq \gamma\leq \mu^*\leq 1 \text{ for } t\geq \tau'_1\right)$. Applying Theorem \ref{thm: estimator concentrations} with $\epsilon(t) = \beta(t) = \beta'_{k^*}(t)$ (this choice is valid for $\epsilon(t)$ due to the definition of $\tau'_1$) and using Claim \ref{claim: error bound}, we have
	\begin{align*}
		\prob(U'_{k^*}(t)\leq \mu^*)
		&\leq \exp\left(- \frac{\beta'_{k^*}(t)^2 t\gamma^2 \left(1-p_V\right)^2}{128M^2 \log^2\left( \frac{2}{\beta'_{k^*}(t)}\right)} \right)
	\end{align*}
	Consider the exponent:
	\begin{align*}
		{\frac{\beta'_{k^*}(t)^2t \gamma^2(1-p_V)^2}{128M^2 \log^2\left(\frac{2}{\beta'_{k^*}(t)}\right)^2}} = { \frac{t\gamma^2(1-p_V)^2}{128M^2}C^2\left( \frac{w\left( \sqrt{\nicefrac{\log t}{t}}\right)}{\log \left(\nicefrac{2}{Cw\left( \sqrt{\frac{\log t}{t}}\right)} \right)}\right)^2 } \geq 2\log t
	\end{align*}
	The final inequality holds for the choice of $C = \frac{32M}{\gamma(1-p_V)}$ and because $\frac{w(x)}{\log\left(\nicefrac{2}{aw(x)}\right)} \geq x$ for $a\geq 1$, as is the case with $C$. Substituting this exponent, completes the proof.
\end{proof}

\begin{lemma}\label{lem: subopt arm concentration}
	Let $\tau'_k$ be as in Equation \ref{eq:Tk'defs}. Then, for all $t\geq \tau'_k$ and $k\neq k^*$, $
	\prob(U'_k(t) >\mu^*) \leq \frac{1}{t^2}.$
\end{lemma}

\begin{proof}
	Note that $Z_k(t)\geq \frac{t}{M}$. Since $w(x)$ is increasing in $x$, using $t\geq \tau'_k$, we can write
	\begin{align*}
		\frac{3}{2}\beta_k(t) + e_k(t) &\leq \frac{3C}{2}w \left( \frac{M\sqrt{ t \log t}}{t}\right) + \frac{\gamma p_V}{2}\\
		&\leq \frac{3C}{2} w\left( \frac{\nicefrac{(\Delta_k - \gamma p_V)}{3C} }{\log\left( \frac{2}{\nicefrac{(\Delta_k - \gamma p_V)}{3C}} \right)} \right) + \frac{\gamma p_V}{2} \\
		&= \frac{3C}{2}\cdot\frac{\Delta_k - \gamma p_V}{3C} + \frac{\gamma p_V}{2} = \frac{\Delta_k}{2} 
	\end{align*}
	
	Here, the penultimate equality follows from the definition of the $w$ function. Therefore, we can write
	\begin{align*}
		\prob \left( U'_{k}(t) > \mu^* \right) &= \prob \left(\tilde{Y}_k(t) > \mu^* - \frac{3 \beta_k(t)}{2} -e_k(t) \right) \\
		&\leq \prob \left(\tilde{Y}_k(t) > \mu^* - \frac{\Delta_k}{2} \right) \\
		&\leq \prob\left(\tilde{Y}_k(t) > \mu_k + \frac{\Delta_k}{2} \right)\\
		&\leq \prob \left(\tilde{Y}_k(t) > \mu_k\left( 1 + \frac{\Delta_k}{2}\right) \right) \\
		&\leq \prob \left(\tilde{Y}_k(t) > \mu_k\left( 1 + \beta'_k(t) \right) \right)
	\end{align*}
	
	The penultimate inequality uses $\mu_k\leq 1$ and the final one follows since $\beta'_k(t)\leq \beta_k(t) \leq \frac{3}{2}\beta(t)_k + e_k(t) \leq \frac{\Delta_k}{2}$.
	Thus, we can now apply Theorem \ref{thm: estimator concentrations} with $\beta(t) = \epsilon(t) = \beta'_k(t)$ and follow the exponent bounding arguments as in Lemma \ref{lem: best arm concentration} to obtain the required result.
\end{proof}

The two lemmas above lead to the following useful corollary:

\begin{corollary}\label{cor: prob of arm play empirical}
	For all suboptimal experts $k\neq k^*$ and $t\geq \tau'_k,$ we have that 
	\begin{align*}
		\prob(k_t = k) \leq \frac{2}{t^2}
	\end{align*}
\end{corollary}
\begin{proof}
	Consider any suboptimal expert $k:\Delta_k>0$ and time $t\geq \tau'_k$. We bound the probability that it is played as follows:
	\begin{align*}
		\prob(k_t= k) &= \prob\left(k_t = k, U'_{k^*}(t)\geq \mu^*\right) + \prob\left(k_t = k, U'_{k^*}(t)<\mu^*\right)\\
		& = \prob\left(k_t = k|U'_{k^*}(t)\geq \mu^*\right)\cdot\prob(U'_{k^*}(t)\geq \mu^*) +\prob\left(k_t = k|U'_{k^*}(t)< \mu^*\right)\cdot\prob(U'_{k^*}(t)< \mu^*)\\
		&\leq \prob\left(k_t = k|U'_{k^*}(t)\geq \mu^*\right) + \prob\left( U'_{k^*}(t)<\mu^*\right)\\
		&\leq \prob(U'_k(t)>\mu^*) + \prob(U'_{k^*}(t)<\mu^*)\\
		&\leq \frac{2}{t^2}.
	\end{align*}
	
	Here, the first inequality uses the fact that probabilities are bounded by 1, then, we use $\{k_t = k|U'_{k^*}(t)\geq \mu^*\}\subseteq \{ U'_k(t)>\mu^*\}$. Finally, we use the results of Lemmas \ref{lem: best arm concentration},\ref{lem: subopt arm concentration} above since $t\geq \tau'_k\geq \tau'_1$.
\end{proof}

\subsection{Proof of Theorem \ref{thm: per-epi regret main paper}}\label{sec: regret theorem proof}
\begin{proof}
	Using Corollary \ref{cor: prob of arm play empirical}, we can bound the regret using the following chain
	\begin{align*}
		\expec[R(T)]&= \sum_{t=1}^T\sum_{k=2}^N \Delta_k\prob(k_t = k)\\
		&\leq {\sum_{t=1}^{\tau'_N-1} \Delta_N + \sum_{t=\tau'_N}^T \Delta_N \prob(k_t = N)} +{ \sum_{t=\tau'_N}^T\sum_{k=2}^{N-1} \Delta_k\prob(k_t = k)}\\
		&\leq { \tau'_N\Delta_N + \Delta_N \frac{\pi^2}{3} + \sum_{t=\tau'_N}^{\tau'_{N-1}-1} \Delta_{N-1} } +{ \sum_{t=\tau'_{N-1}}^T \Delta_{N-1} \prob(k_t = N-1)} +{ \sum_{t=\tau'_{N-1}}^T\sum_{k=2}^{N-2} \Delta_k\prob(k_t = k)}\\
		&\leq {\Delta_N\left( \tau'_N + \frac{\pi^2}{3}\right) + \Delta_{N-1} \left( (\tau'_{N-1}-\tau'_N)+ \frac{\pi^2}{3}\right) + ...} +{\Delta_2\left( (\tau'_2 - \tau'_3) + \frac{\pi^2}{3}\right)}\\
		&= \frac{\pi^2}{3} \sum_{k=2}^N \Delta_k + \tau'_N\Delta_N + \sum_{k=2}^{N-1} \left( \tau'_k - \tau'_{k+1}\right)
	\end{align*}
\end{proof}

\section{Proofs of results in Section \ref{sec: full info regret}}

All our results for ED-UCB can be used to derive results for D-UCB by setting $\xi = 0$ and replacing the empirical IS ratios and divergences with their true values. In particular, Theorem \ref{thm: estimator concentrations} can be modified to prove Theorem \ref{lem:clipped}. Then, with $\tau_i$ defined as in Equation \ref{eq:Tk defs}, we arrive at high probability error bounds for the UCB indices of the best and suboptimal arms separately (analogous to Lemmas \ref{lem: best arm concentration}, \ref{lem: subopt arm concentration}). These are then used to conclude Corollary \ref{cor: prob of arm play}. The result of Theorem \ref{thm: one episode regret} then follows from arguments similar to those in Theorem \ref{thm: per-epi regret main paper}. We present short versions of these results below.

\subsection{Estimator Concentrations}

We begin with the following counterpart to Remark \ref{rem: intermediate bound for mu}.

\begin{lemma}
	\label{lem:mean}
	For all times $l$ and all experts $j\in [N]$,  we have,
	\begin{align}
		\mu_j - \frac{\epsilon(t)}{2} \leq \mu_j(l) \leq \mu_j \label{eq:prev_paper}.
	\end{align}
	where $\mu_j(l)$ is defined in the proof of Theorem~\ref{lem:clipped}.
\end{lemma}

\begin{proof}
	We first note that under the filtration $\mathcal{F}_{l-1}$, $M_{jk(l)}$ is a constant and $k(l)$ is fixed. Therefore, following the notation in Theorem \ref{lem:clipped},  we have the following chain,
	\begin{align*}
		&\mu_j(l) = \expec \left[L_j(l) \vert \mathcal{F}_{l-1} \right] \\
		&~~~~= \expec_{k(l)} \left[ Y_l\frac{\pi_j(V_{k(l)}(l) \vert X_{k(l)}(l))}{\pi_{k(l)}(V_{k(l)}(l) \vert X_{k(l)}(l))}\right]  \\
		&~~~~~~~~~~~~-\expec_{k(l)} \left[Y_l \frac{\pi_j(V_{k(l)}(l) \vert X_{k(l)}(l))}{\pi_{k(l)}(V_{k(l)}(l) \vert X_{k(l)}(l))}\times\ind\left\{ r_l > \alpha_l\right\} \right] \\
		&~~~~\stackrel{(i)}{\geq} \mu_j - \prob_{j}\left( \frac{\pi_j(V_{k(l)}(l) \vert X_{k(l)}(l))}{\pi_{k(l)}(V_{k(l)}(l) \vert X_{k(l)}(l))}  > 2\log(2/\epsilon (t))M_{jk(l)} \right) \\
		&~~~~\stackrel{(ii)}{\geq} \mu_j - \frac{\epsilon(t)}{2}. 
	\end{align*}
	
	Here, (i) follows from the fact that $Y \in [0,1]$ and (ii) follows from Lemma 2 in~\cite{sen2017identifying}. 
\end{proof}

Now, we are ready to prove Theorem \ref{lem:clipped}.

\begin{proof}[Proof of Theorem \ref{lem:clipped}]
	
	We will reuse notation from Theorem \ref{thm: estimator concentrations} for convenience. Let $\mathcal{F}_s$ be the filtration formed by the observation until time $s$ \textit{and} the expert chosen at time $s+1$. We define the martingale $\{A_s\}$ with $A_0=0$ and 
	
	\begin{align*}
		A_s = \sum_{l=1}^{s} \frac{L_j(l)}{M_{jk(l)}} -  \sum_{l=1}^{s} \expec\left[ \frac{L_j(l)}{M_{jk(l)}} \bigg \vert \mathcal{F}_{l-1} \right]. 
	\end{align*}
	
	where to ease notation, we define 
	\begin{align*}
		r_l := \frac{\pi_j(V_{k(l)}(l) \vert X_{k(l)}(l))}{\pi_{k(l)}(V_{k(l)}(l) \vert X_{k(l)}(l))}, 
		~~~~\alpha_l := 2\log(2/\epsilon(t))M_{jk(l)}, 
		~~~~L_j(l) :=  Y_l \times r_l \times\ind\left\{ r_l \leq \alpha_l \right\}.
	\end{align*}
	
	With $\mu_j(l) := \expec[L_j(l) | F_{l-1}]$ and $B_t = \sum_{l = 1}^t \frac{L_j(l)}{M_{jk(l)}}$, we rewrite $A_s$ as $A_s = B_s -  \sum_{l=1}^{s} \frac{\mu_j(l)}{M_{jk(l)}}.$
	
	Since $|A_s - A_{s-1}| \leq 4\log(2/\epsilon(t))$, the Azuma-Hoeffding inequality implies that
	\begin{align}
		\prob\left( \bigg \lvert B_t -  \sum_{l=1}^{t} \frac{\mu_j(l)}{M_{jk(l)}} \bigg \rvert \geq \chi\right) \leq 2\exp \left( - \frac{\chi^2}{32 t (\log (2 / \epsilon(t)))^2 }\right). \label{eq:martingale}
	\end{align}	
	
	\noindent\textbf{1. Upper tail} We have that 
	\begin{align*}
		\prob\left(  B_t \geq \sum_{l=1}^{t} \frac{\mu_j(l)}{M_{jk(l)}} + \chi \right) \leq \exp \left( - \frac{\chi^2}{32t (\log (2 / \epsilon(t)))^2 }\right)
	\end{align*}
	Consider the following chain:
	\begin{align*}
		(1+\beta(t))\left( \max_{l\in[t]} \mu_j(l)\right) \sum_{l = 1}^t \frac{1}{M_{jk(l)}} \geq \sum_{l = 1}^t \frac{\mu_j(l)}{M_{jk(l)}}  + \frac{\beta(t) t}{M}\max_{l\in [t]} \mu_j(l) \geq \sum_{l = 1}^t \frac{\mu_j(l)}{M_{jk(l)}}  + \frac{\gamma\beta(t) t}{2M}
	\end{align*}
	
	where the final inequality comes about as a consequence of $\epsilon(t)<\gamma$ and $\mu_j(l) \geq \mu_j -\frac{\epsilon(t)}{2}\geq \gamma - \frac{\epsilon(t)}{2}$. The latter is proved in Lemma \ref{lem:mean}. We also use the fact that $M := \max_{j,k} M_{jk}$. Thus, we have 
	\begin{align*}
		\prob\left(  B_t \geq (1 + \beta(t) ) \left(\max_l {\mu_j(l)} \right)\sum_{l=1}^{t} \frac{1}{M_{jk(l)}}  \right) &\leq \prob \left(  B_t \geq \sum_{l=1}^{t} \frac{\mu_j(l)}{M_{jk(l)}} + \frac{\gamma\beta(t) t}{2M} \right) \\
        &\leq \exp \left( - \frac{\gamma^2\beta(t)^2t}{128 M^2 (\log (2 / \epsilon(t)))^2 }\right). 
	\end{align*}
	This implies that,
	\begin{align}
		\prob \left( \hat{\mu}_k(t) \geq (1 + \beta(t) ) (\max_l {\mu_j(l)} ) \right) \leq \exp \left( - \frac{\gamma^2\beta(t)^2t}{128 M^2 (\log (2 / \epsilon(t)))^2 }\right). \label{eq:upper_tail}
	\end{align}
	
	\noindent\textbf{2. Lower Tail} We have that 
	\begin{align*}
		\prob \left(  B_t \leq \sum_{l=1}^{t} \frac{\mu_j(l)}{M_{jk(l)}} - \chi \right) \leq \exp \left( - \frac{\chi^2}{32t (\log (2 / \epsilon(t)))^2 }\right)
	\end{align*}
	
	Using similar arguments as above, we can write
	
	\begin{align*}
		\prob \left(  B_t \leq (1 - \beta(t) ) (\min_l {\mu_j(l)} )\sum_{l=1}^{t} \frac{1}{M_{jk(l)}}  \right) &\leq \prob \left(  B_t \leq \sum_{l=1}^{t} \frac{\mu_j(l)}{M_{jk(l)}} - \frac{\gamma\beta(t) t}{2M} \right) \\
        &\leq \exp \left( - \frac{\gamma^2\beta(t)^2t}{128 M^2 (\log (2 / \epsilon(t)))^2 }\right). 
	\end{align*}
	
	This implies that,
	\begin{align}
		\prob \left( \hat{\mu}_k(t) \leq (1 - \beta(t) ) (\min_l {\mu_j(l)} ) \right) \leq \exp \left( - \frac{\gamma^2\beta(t)^2t}{128 M^2 (\log (2 / \epsilon(t)))^2 }\right). \label{eq:lower_tail}
	\end{align}
	Since the choice of $j\in[N]$ was arbitrary, Combining Equation~\ref{eq:upper_tail}, ~\ref{eq:lower_tail} above with Equation \ref{eq:prev_paper} from the lemma above, we have the result. 
	
\end{proof}

\subsection{Per-expert concentrations}
We begin with the best expert and show that it is overestimated with high probability.

\begin{lemma}\label{lem:ubound} 
	Let $\tau_1 = \min\left\{ t: \beta'(t) := Cw\left(\frac{\sqrt{c_1t\log t}}{t} \right)\leq \gamma\right\}$ be as in Equation \ref{eq:Tk defs}. Then, for all $t\geq \tau_1$, the index of the best arm formed using the Clipped Estimator satisfies
	\begin{align*}
		\prob \left( U_{k^*}(t) > \mu^* \right) \geq 1 - \frac{1}{t^2}.
	\end{align*}
\end{lemma}

\begin{proof}
	Since $Z_k(t)\leq t$ and $w(x)$ is increasing, we have that $\beta(t)\geq \beta'(t)$. Now, we have the following chain,
	\begin{align*}
		\PP \left( U_{k^*}(t) \leq \mu^* \right) &= \PP \left( \hat{\mu}_{k^*}(t) \leq \mu^* - \frac{3}{2}C w \left(\frac{\sqrt{c_1 t \log t}}{Z_k(t)}\right) \right) \\
		& \stackrel{(i)}{\leq} \PP \left( \hat{\mu}_{k^*}(t) \leq \mu^* -\mu^*C w \left(\frac{\sqrt{c_1 t \log t}}{t}\right) - \frac{1}{2}C w \left(\frac{ \sqrt{c_1 t \log t}}{t}\right) \right) \\
		&\stackrel{(ii)}{\leq} \PP \left( \hat{\mu}_{k^*}(t) \leq \mu^* -\mu^* \beta'(t) - (1 - \beta'(t)) \frac{1}{2} \beta'(t) \right) \\
		&\leq \PP\left(\hat\mu_{k^*}(t) \leq (1-\beta'(t))\left(\mu^* - \frac{\beta'(t)}{2}\right)  \right)
	\end{align*}
	
	Here, (i) uses the observation above and that $\mu^* \leq 1$. (ii) follows from the fact that $\beta'(t) \leq 1$. Now, we use Theorem \ref{lem:clipped} with $\beta(t)$ set to be sample path independent $\beta'(t)$ to write 
	\begin{align*}
		\PP \left( U_{k^*}(t) \leq \mu^* \right) &\leq \exp\left( -\frac{\gamma^2 \beta'(t)^2 t}{128M^2 (\log(2/\beta'(t)))^2} \right)
	\end{align*}
	
	Similar to the arguments in Lemma \ref{lem: best arm concentration}, with $C = \frac{16M}{\gamma}$, we can upper bound the exponent by $-2\log t$. This leads to the result.
\end{proof}

We now bound the probability of overestimating a suboptimal expert.

\begin{lemma}
	\label{lem:lbound}
	
	Let $\tau_1$ be as in Lemma \ref{lem:ubound} and $\tau_k$ as in Equation \ref{eq:Tk defs}. Then, for any $k\neq k^*$, for any $t \geq \tau_k$, we have that 
	\begin{align*}
		\PP \left( U_{k}(t) < \mu^* \right) \geq 1 - \frac{1}{t^2}.
	\end{align*}
\end{lemma}

\begin{proof}
	We have that $Z_k(t)\geq \frac{t}{M}$ and $t\geq\frac{9C^2 M^2 \log T \log^2\left( \nicefrac{6C}{\Delta_k}  \right)}{\Delta_k^2}.$ Additionally, $w(x)$ is increasing in $x$. Therefore, we can write that
	\begin{align*}
		\frac{3C}{2}w \left( \frac{\sqrt{c_1 t \log t}}{Z_k(t)}\right) &\leq \frac{3C}{2}w \left( \frac{M\sqrt{c_1 t \log t}}{t}\right)\\
		&\leq \frac{3C}{2}w \left( \frac{M\sqrt{c_1 t \log T}}{t}\right) \nonumber \\
		&\leq \frac{3C}{2}w \left( \frac{M\sqrt{c_1  \log T}}{\frac{3CM\sqrt{c_1\log T} \log \left(\nicefrac{6C}{\Delta_k} \right)}{\Delta_k}}\right)\\
		&= \frac{3C}{2} w\left( \frac{\nicefrac{\Delta_k}{3C} }{\log\left(  \frac{2}{\nicefrac{\Delta_k}{3C}}  \right)} \right)\\
		&= \frac{\Delta_k}{2} 
	\end{align*}
	Where the last equality follows from the fact that $w(x) = y \iff \frac{y}{\log\left(\nicefrac{2}{y}\right)} = x$. Using the argument above along with the fact that $\Delta_k = \mu^* - \mu_k$ and $\mu_k<\mu^*\leq 1$, we have the following chain:
	
	\begin{align*}\label{eq:lboundeq}
		\PP \left( U_{k}(t) > \mu^* \right) &= \PP \left(\hat{\mu}_k(t) > \mu^* - \frac{3 \beta(t)}{2} \right) \\
		&\leq \PP \left(\hat{\mu}_k(t) > \mu^* - \frac{\Delta_k}{2}\right) \\
		&\leq  \PP\left(\hat\mu_k(t) > \mu_k + \frac{\Delta_k}{2} \right)\\
		&\leq \PP \left(\hat{\mu}_k(t) > \mu_k\left( 1 + \frac{\Delta_k}{2}\right) \right) \\
		& \stackrel{(i)}{\leq} \PP \left(\hat{\mu}_k(t) > \mu_k\left( 1 + \frac{3C}{2}w \left( \frac{M \sqrt{t \log t}}{t}\right) \right) \right) \\
		&\leq \exp\left(-\frac{\gamma^2 \left(\nicefrac{3C}{2}\right)^2 t }{128M^2} \times \left( \frac{w\left(M\sqrt{\nicefrac{\log t}{t}}\right)}{\log \left(\nicefrac{2}{Cw\left(\sqrt{\nicefrac{\log t}{t}} \right) } \right)} \right)^2\right)\numberthis
	\end{align*}
	
	In (i) we have used the fact that $\Delta_k/2 \geq \frac{3C}{2}w \left( \frac{M \sqrt{t \log t}}{t}\right)$ from the chain just before. Here, the final inequality applies Theorem~\ref{lem:clipped} (bounds for the upper tail error) with $\chi(t) = \frac{3C}{2}w \left( \frac{M\sqrt{ t \log t}}{t}\right)$ and $\epsilon(t) = \beta'(t)$ defined in Lemma~\ref{lem:ubound}. Upper bounding the exponent by $-2\log t$ using arguments in Lemma \ref{lem: best arm concentration} gives us the result.
	
\end{proof}

These lead to Corollary \ref{cor: prob of arm play}.

\begin{proof}[Proof of Corollary \ref{cor: prob of arm play}]
	The proof follows the same chain of reasoning as Corollary \ref{cor: prob of arm play empirical} with all the empirical quantities replaced by their full-information counterparts. We have 
	\begin{align*}
		\prob(k_t= k) &= \prob\left(k_t = k, U_{k^*}(t)\geq \mu^*\right) + \prob\left(k_t = k, U_{k^*}(t)<\mu^*\right)\\
		&\leq \prob\left(k_t = k|U_{k^*}(t)\geq \mu^*\right) + \prob\left( U_{k^*}(t)<\mu^*\right)\\
		&\leq \prob(U_k(t)>\mu^*) + \prob(U_{k^*}(t)<\mu^*)\\
		&\leq \frac{2}{t^2}.
	\end{align*}
	
	Again, we use that $\{k_t = k | U_{k^*}(t)\geq \mu^* \}\subseteq \{U_k(t)\geq \mu^* \}$ in the second inequality. The result follows by applying Lemmas \ref{lem:ubound}, \ref{lem:lbound}.
\end{proof}

\subsection{Proof of Theorem \ref{thm: one episode regret}}

\begin{proof}
	Using Corollary \ref{cor: prob of arm play}, we can bound the regret using the following chain
	\begin{align*}
		\expec[R(T)]&= \sum_{t=1}^T\sum_{k=2}^N \Delta_k\prob(k_t = k)\\
		&\leq {\sum_{t=1}^{\tau_{N}-1} \Delta_N + \sum_{t=\tau_N}^T \Delta_N \prob(k_t = N)}+{\sum_{t=\tau_N}^T\sum_{k=2}^{N-1} \Delta_k\prob(k_t = k)}\\
		&\leq { \tau_N\Delta_N + \Delta_N \frac{\pi^2}{3} +\sum_{t=\tau_N}^{\tau_{N-1}-1} \Delta_{N-1} } +{\sum_{t=\tau_{N-1}}^T \Delta_{N-1} \prob(k_t = N-1)} +{\sum_{t=\tau_{N-1}}^T\sum_{k=2}^{N-2} \Delta_k\prob(k_t = k)}\\
		&\leq {\Delta_N\left( \tau_N + \frac{\pi^2}{3}\right) + \Delta_{N-1} \left( (\tau_{N-1}-\tau_N)+ \frac{\pi^2}{3}\right) + ...} +{\Delta_2\left( (\tau_2 - \tau_3) + \frac{\pi^2}{3}\right)}\\
		&= \frac{\pi^2}{3} \sum_{k=2}^N \Delta_k + \tau_N\Delta_N + \sum_{k=2}^{N-1} \left( \tau_k - \tau_{k+1}\right)
	\end{align*}
\end{proof}

\section{Proofs of Results in Section \ref{sec: improved scaling} and Tighter Regret Bounds}\label{app: improved scaling}

We prove our improved computational complexity result:

\begin{proof}[Proof of Lemma \ref{lem: improved computation}]
	Let $S(t) = S'(t)\cup k_t$ for $S'(t)$ defined as in Algorithm \ref{alg:DUCB-lite}. Define $E_k(t) = \{ \exists t'\in[t-\tau,t]: k\in S(t')\}$ as the event that expert $k$ has been updated at least once in the last $\tau$ time steps from $t$, for some $\tau$ to be chosen. We have that 
	\begin{align*}
		\prob(E_k^C(t)) &= \prob(\not\exists t'\in[t-\tau,t]:k\in S(t')) \\
            &\leq \prob(\not\exists t'\in[t-\tau,t]: k\in S'(t)) \\
            &\leq \left(1- \frac{\log N}{N-1}\right)^\tau \\
            &\leq \left(1-\frac{\log N}{N}\right)^\tau.
	\end{align*}
	Recall that in vanilla D-UCB, we have that for $t\geq \tau_1$, $\prob(U_k^*(t) \leq \mu^*)\leq t^{-2}$(analog to Lemma \ref{lem: best arm concentration}). In case of D-UCB-lite, we can write for $t\geq \tau_{1,\beta}$,
	\begin{align*}
		\prob(B_k^*(t)\leq \mu^*) &= \prob(B_k^*(t)\leq \mu^*, E_{k^*}(t))\\
		&~~~~+ \prob(B_k^*(t)\leq \mu^*, E_{k^*}^C(t))\\
		&\leq \prob(B_k^*(t)\leq \mu^*|E_{k^*}(t))+\prob(E_{k^*}^C(t))\\
		&= \prob(U_{k^*}(\lfloor t'^{\frac{1}{\beta}}\rfloor)\leq\mu^*) + \prob(E_{k^*}^C(t))\\
		&~~~~~~~~~~(\text{for some $t'\in[t-\tau,t]$})\\
		&\leq \frac{1}{t'^{\frac{2}{\beta}}} +  \left(1-\frac{\log N}{N}\right)^\tau\\
		&\leq \frac{1}{(t-\tau)^\frac{2}{\beta}} +  \left(1-\frac{\log N}{N}\right)^\tau\\
		&= \frac{1}{t^2} + \frac{1}{(t-2c\log t)^\frac{2}{\beta}}.
	\end{align*}
	Where the final inequality holds for the choice of $\tau = 2c\log t$. Similarly, we for any suboptimal arm $k$, after time $\tau_{k,\beta}$, we can write $\prob(B_k(t) >\mu^*) \leq t^{-2} + (t-2c\log t)^{-\frac{2}{\beta}}$. 
	
	The regret result follows using these two statements and the regret decomposition in the proof of our regret theorem in Appendix \ref{sec: regret theorem proof}.
\end{proof}

The result of Theorem \ref{thm: one episode regret} implies that average regret of D-UCB is bounded by a problem-dependent constant (for any $i\in[N], \tau_i$ is a constant). However, there are two things to note here: First, the upper bound expression is linear in the number of experts $N$ and second, the algorithm updates each of these $N$ experts once per iteration. In this section, we further examine these observations.

\noindent\textbf{Scaling with $N$: }We first provide the regret improvement result in the case where the suboptimality gaps are drawn from a generative model. Specifically, consider a generative model where $\Delta_{3}\leq...\leq\Delta_{N}$ are the order statistics of $N-2$ random variables drawn i.i.d uniform over the interval $[\Delta_{2},1]$. Let $p_{\Delta}$ denote the measure over these $\Delta$'s.

\begin{corollary}\label{cor:logNregret}
Under the generative model $p_\Delta$ above, the regret of Algorithm \ref{alg:DUCB} can be bounded as $\EE_{p_{\Delta}} \left[ R(T)\right] = \mathcal{O} \left(\frac{M^4 \log N \log^2(1/\Delta_{(2)}) }{\Delta_{(2)}} \right)$.
\end{corollary}

\begin{proof}[Proof of Corollary \ref{cor:logNregret}]
	The analog to Lemma \ref{lem: subopt arm concentration} in the full information setting guarantees that for arm $k$, $\prob(U_k(t) \geq \mu^*) \leq t^{-2}$ for $t> \frac{\alpha C^2M^2\log^2\left( \frac{6C}{\Delta_2}\right)}{\Delta_k^2}$ for an $\alpha$ large enough. Note here $\Delta_k$ in the log term has been replaced with a smaller $\Delta_2$. The regret can now be decomposed as
	\begin{align*}
		R(T) &\leq \frac{\pi^2}{3}\sum_{k=2}^N \Delta_k + \frac{\alpha C^2M^2\log^2\left( \frac{6C}{\Delta_2}\right)}{\Delta_N} + \sum_{k=2}^{N-1} \frac{\alpha C^2M^2\log^2\left( \frac{6C}{\Delta_2}\right)}{\Delta_k} \left(1 - \frac{\Delta_k^2}{\Delta_{k+1}^2}\right)\\
		&\leq \mathcal{O}\left( \frac{ M^4\log^2\left( \frac{1}{\Delta_2}\right)}{\Delta_2}\left( 1 + \sum_{k=2}^{N-1} 1 - \frac{\Delta_k^2}{\Delta_{k+1}^2}\right)\right).
	\end{align*}
	
	It only remains to prove that under the considered generative model, the inner expression is $\mathcal{O}(\log N)$. We start by observing that by Jensen's inequality, we get 
	
	\begin{align*}
		1- \expec_{p_\Delta}\left[\frac{\Delta_k^2}{\Delta_{k+1}^2}\right] &\leq 1- \expec_{p_\Delta}\left[\frac{\Delta_k}{\Delta_{k+1}}\right]^2.
	\end{align*}
	
	Let $X = \Delta_k, Y = \Delta_{k+1}$ for some $k\geq 3$. Then, we have that the joint distribution of $X,Y$ under the generative model to be 
	\begin{align*}
		f(x,y) = {\frac{(N-1)!}{(k-1)!(N-k-3)!} \left(\frac{x - \Delta_2}{1-\Delta_2}\right)^{k-1}\left(1-\frac{y - \Delta_2}{1-\Delta_2}\right)^{N-3-k}\cdot\frac{1}{(1-\Delta_2)^2}.}
	\end{align*}
	Thus, we have 
	\begin{align*}
		\expec[X/Y] &= \int_{y=\Delta_2}^1\int_{x=\Delta_2}^y \left(\frac{x}{y}\frac{(N-1)!}{(k-1)!(N-k-3)!} \left(\frac{x - \Delta_2}{1-\Delta_2}\right)^{k-1}\right.\\
        &~~~~~~~~\cdot{\left.\left(1-\frac{y - \Delta_2}{1-\Delta_2}\right)^{N-3-k}\cdot\frac{1}{(1-\Delta_2)^2}dxdy\right)}\\
		&={\int_{b=0}^1\int_{a=0}^b\left( \frac{(1-\Delta_2)a + \Delta_2}{(1-\Delta_2)b + \Delta_2}\cdot \frac{(N-1)!}{(k-1)!(N-k-3)!}\right.} \cdot{\left.a^{k-1}b^{N-3-k}dxdy\right)}\\
		&\leq{\int_{b=0}^1\int_{a=0}^b\left( \frac{a}{b}\cdot \frac{(N-1)!}{(k-1)!(N-k-3)!}a^{k-1}b^{N-3-k}dxdy\right)}\\
		&=\frac{k}{k+1}
	\end{align*}
	Therefore, we have 
	\begin{align*}
		\expec_{p_\Delta}\left[ \sum_{k=2}^{N-1} 1- \frac{\Delta_k^2}{\Delta_{k+1}^2}\right] &\leq 1+\sum_{k=3}^{N-1} 1 - \frac{k^2}{(k+1)^2} = 1 + \sum_{k=3}^{N-1}\frac{2k+1}{(k+1)^2} \leq 1 + \sum_{k=3}^{N-1}\frac{2}{(k+1)} \leq 1+ 2\log N
	\end{align*}
	This completes the proof.
\end{proof}

Note that this Corollary shows that under the defined generative model for the gaps in the means of experts, the dependence on $N$ can be improved from linear to \textit{logarithmic}.

\section{Proofs for Results in Section \ref{sec: episodic setting}}
We begin by proving that with $A, n$ as defined, with high probability if each expert is played $A$ times, each context $x\in\calx$ is seen $n$ times. 

\begin{proof}[Proof of Lemma \ref{lem: samples lemma}]
	We fix a context $x$ and expert $i$ arbitrary. Then, we have that
	\begin{align*}
		\prob(x \text{ seen $< n $ times after $A$ pulls of } i) &\leq \prob(Z\leq n) \\
        &\leq \exp\left(-2A\left(p(x) - \frac{n}{A}\right)^2\right) \\
        &\leq \exp\left( {\scriptstyle -2A\left( p_X^2 - 2\frac{np_X}{A} \right)}\right) = \frac{1}{|\calx|NT\sqrt{E}}.
	\end{align*}
	In the above, $Z\sim Binomial(A,p(x))$ and the final inequality uses the definition of $A$. We say expert $i$ is incomplete if there exists a context $x\in\calx$ s.t. $x$ has been seen $<n$ times after $A$ pulls of arm $i$. Then, a series of union bounds gives us the result.
	\begin{align*}
		\prob(\mathcal{E}^C) &= \prob(\text{$\exists i\in[N]$ incomplete}) \\
            &\leq \sum_{i=1}^N \sum_{x\in\calx} \prob(\text{$x$ seen $<n$ times in $A$ pulls of arm } i) \\
            &\leq \sum_{i=1}^N \sum_{x\in\calx} \frac{1}{|\calx|NT\sqrt{E}} = \frac{1}{T\sqrt{E}}
	\end{align*}
\end{proof}

Next, we provide the final regret result for the agent bootstrapped with $A$ samples per expert. 

\begin{proof}[Proof of Theorem \ref{thm: regret with bootstrapping}]
	Lemma \ref{lem: samples lemma} gives us that with probability at least $1-\frac{1}{T\sqrt{E}}$, each expert has at least $n$ samples before the agent interacts with the environment. Under this event, using Lemma \ref{lem: empirical probability concentrations}, the agent can build empirical experts with maximum error $\xi$ w.p. at least $1-T^{-1}$. Therefore, conditioning on the event $\mathcal{E}$ and then on the event that the approximate experts are accurate, we have the following chain:
	\begin{align*}
		R(E,T) &= \sum_{e\in[E]} R_e(T) \leq { \frac{1}{T\sqrt{E}}\times ET + \left(1-\frac{1}{T\sqrt{E}}\right)\left(\frac{1}{T}\times ET + \left(1-\frac{1}{T}\right) \sum_{e=1}^E R(\Delta_e)\right)}\\
		&\leq \sqrt{E} + E + \left(\sum_{e=1}^T R(\Delta_e)\right)\left(1 +\frac{1}{T^2\sqrt{E}} \right).
	\end{align*} 
\end{proof}

\section{Additional Empirical Evaluations}\label{sec: additional exp}
In this section, we will provide a few additional experimental results that shed light on some of the key assumptions we use to develop D-UCB and ED-UCB. First, we will see how the precision of the empirical policy estimates affects the regret of ED-UCB and then move on to the assumption of bounded divergence between experts.

\subsection{Precision of Empirical Estimates}

\begin{figure}[t]
    \centering
    \begin{subfigure}[b]{0.45\textwidth}
        \includegraphics[width=\textwidth]{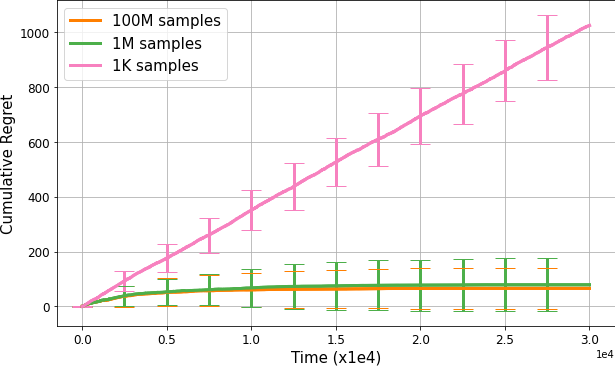}
        \caption{Full-size figure}
    \label{fig: precision full size}
    \end{subfigure}%
    \begin{subfigure}[b]{0.45\textwidth}
        \includegraphics[width=\textwidth]{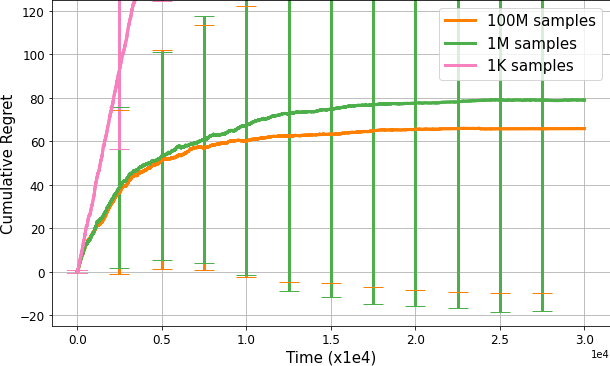}
        \caption{Zoomed in}
    \label{fig: precision zoom}
    \end{subfigure}
    \caption{Precision of Empirical Estimates on regret of ED-UCB: The experiment consists of one episode of $3\times10^4$ steps. The legend indicates the number of samples used to form the empirical expert policies used by ED-UCB in Algorithm \ref{alg:EDUCB}. Plots are averaged over 300 independent runs. The results suggest that using estimates with higher precision leads to lower regret.}
    \label{fig: precision of empirical estimates}
\end{figure}

For the results shown in Figure \ref{fig: precision of empirical estimates}, we consider the same setting as our image classification setup in Section \ref{sec: experiments}. Instead of using Theorem \ref{thm: regret with bootstrapping} to instruct the number of samples to be used to form the empirical expert policies, we set a fixed budget of $100M, 1M$ and $1K$ samples. Then, we split these samples across the different contexts using a fixed context distribution. We reuse this same distribution in the online evaluation. We then spawn three instances of ED-UCB with $\xi$ being set at the value recommended by \ref{thm: per-epi regret main paper} which is inaccurate for all three sets of empirical estimates. All other parameters are kept unchanged. As is to be expected, the results show that there is a inverse relationship between the number of samples used to compute the empirical policies (or equivalently, their precision) and the regret they achieve. 

However, we note the following: We believe that our recommendation for the number of samples to use in Lemma \ref{lem: samples lemma} is loose. The constant regret achieved using 1M samples in Figure \ref{fig: precision of empirical estimates} provides potential evidence of this. However, we stress that the results displayed in the figure are only a one random sample path (not cherry-picked) of choosing a random context distribution, partitioning the samples randomly across contexts and sampling random rewards according to this distribution. That is, the figure by itself is not proof that using lesser samples than recommended can guarantee constant regret.

\begin{figure}[t]
    \centering
    \begin{subfigure}[b]{0.45\textwidth}
        \includegraphics[width=\textwidth]{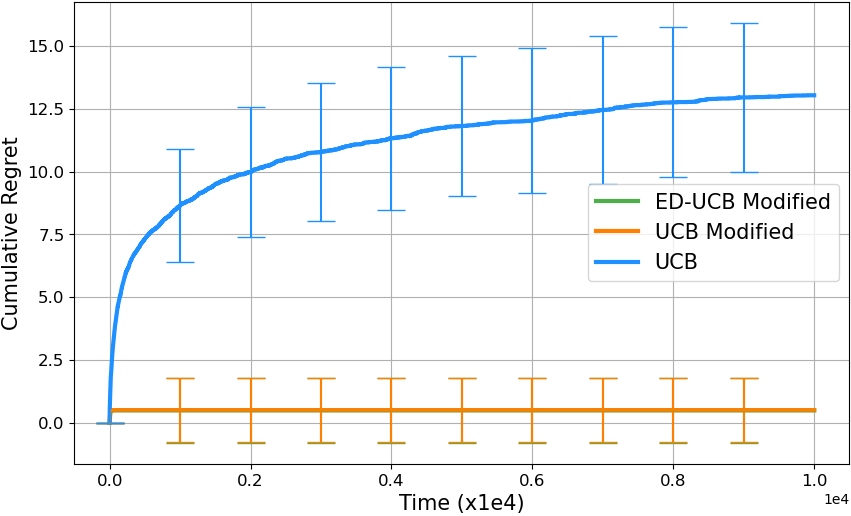}
        \caption{Regret under Reward Setting 1}
    \label{fig: unbounded full size A}
    \end{subfigure}%

    \begin{subfigure}[b]{0.45\textwidth}
        \includegraphics[width=\textwidth]{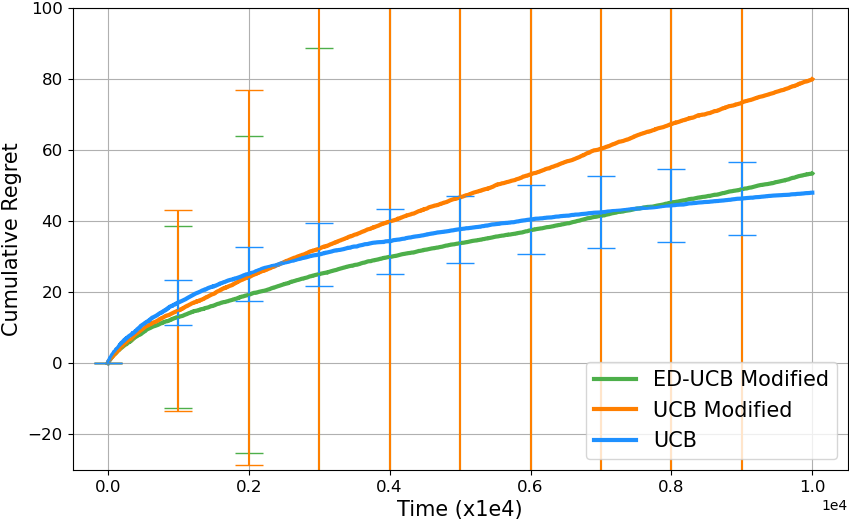}
        \caption{Regret under Reward Setting 2}
    \label{fig: unbounded full size B}
    \end{subfigure}
    \caption{Unbounded Divergence with Modified D-UCB and ED-UCB: The experiments consist of one episode of $3\times10^3$ steps. Plots are averaged over 250 independent runs. We consider modified versions of our proposed algorithms and the toy environments with setups detailed in Section \ref{sec: unbounded M} where the maximal divergence between a pair of experts is unbounded. In the first setting, the mean reward and the probability of picking the problematic arm are low. Therefore, it does not affect the regret much and thus we only suffer constant regret as before. However, in the second setting, the low-probability problematic arm has high mean reward and thus leading to logarithmic exploration much like vanilla UCB.}
    \label{fig: unbounded M}
\end{figure}
\subsection{Infinite Divergence}\label{sec: unbounded  M}
Here, we consider the case of unbounded divergence where our D-UCB and ED-UCB algorithms are not suitable. In this case, the $M_ij$ measure is infinite for some tuple of experts $i,j$. This occurs when there exists an arm that is never recommended by expert $j$. To fit this, we create modified versions of the two algorithms above. Specifically, we replace $D_{f_1}(\pi_i \Vert\pi_j),M_{ij}$ in D-UCB and $\underline{D}_{f_1}(i\Vert j), \underline{M}_{ij}$ in ED-UCB with
\begin{align*}
    D'_{f_1}(\pi_i\Vert\pi_j) &= \sum_{x\in\calx} p(x) \sum_{v\in\calv:\pi_j(v|x)>0} f\left( \frac{\pi_i(v|x)}{\pi_j(v|x)}\right)\pi_j(v|x), &M'_{ij} &= 1+\log(1+D'_{f_1}(\pi_i\Vert \pi_j))\\
\underline{D}'_{f_1}(i||j) &= p_X \sum_{x\in\calx}\sum_{v\in\calv:\hat\pi_j(v|x)>0} \left(\hat{\pi}_j(V|X) - \xi\right)f_1\left(\undr_{ij}(V|X)\right), &\underline{M}'_{ij} &= 1 + \log(1 + \underline{D}'_{f_1}(i||j))
\end{align*}

The above quantities are finite surrogates for the potentially infinite original divergence measures. They are constructed by ignoring the terms from the problematic arm. We note here that these modified versions are only sensible when the zero-probability arm is chosen with low probability. This modification makes it so that the Clipped importance sampling estimates are always well defined. Further, samples of some arm $v\in\calv$ that is only picked by expert $k\in[N]$ are only used to update the estimates of this expert and have no impact on any other estimates. This also leads to unequal number of samples being used to compute the estimates of different experts. 

For the experiments, we create toy environments with 2 contexts $X_1, X_2$ and 3 actions $A_1, A_2, A_3$. We consider 2 experts operating on these environments with expert policies and reward settings summarized in Tables \ref{tab: expert policies} and \ref{tab: reward distributions}.

\begin{table}[h]
\caption{Expert Policies}
\label{tab: expert policies}
\begin{tabular}{|l|l|l|l|l|l|l|l|l|}
\cline{1-4} \cline{6-9}
Expert 1 & $A_1$ & $A_2$ & $A_3$ &  & Expert 2 & $A_1$ & $A_2$ & $A_3$ \\ \cline{1-4} \cline{6-9} 
$X_1$    & 0.8   & 0.1   & 0.1   &  & $X_1$    & 0.2   & 0.8   & 0.0   \\ \cline{1-4} \cline{6-9} 
$X_2$    & 0.2   & 0.8   & 0.1   &  & $X_2$    & 0.8   & 0.2   & 0.0   \\ \cline{1-4} \cline{6-9} 
\end{tabular}
\end{table}

\begin{table}[h]
\caption{Reward Distributions}
\label{tab: reward distributions}
\begin{tabular}{|l|l|l|l|l|l|l|l|l|}
\cline{1-4} \cline{6-9}
Reward Setting 1 & $A_1$ & $A_2$ & $A_3$ &  & Reward Setting 2 & $A_1$ & $A_2$ & $A_3$ \\ \cline{1-4} \cline{6-9} 
$X_1$    & 0.9   & 0.1   & 0.1   &  & $X_1$    & 0.1   & 0.1   & 0.9   \\ \cline{1-4} \cline{6-9} 
$X_2$    & 0.1   & 0.9   & 0.1   &  & $X_2$    & 0.1   & 0.1   & 0.9   \\ \cline{1-4} \cline{6-9} 
\end{tabular}
\end{table}

As Expert 2 never picks arm $A_3$, the upper bound on the divergence is no longer finite. For context distribution, we pick $X_1$ with probability 0.55 and $X_2$ with probability 0.45. We set $p_X=0.1, p_V=0.05$ and note that the value of $p_V$ is inaccurate (the true value is 0). In Figure \ref{fig: unbounded M}, we present the results of our experiments on both reward Settings. Under Setting 1, the rewards from $A_3$ as well as the chance of Expert $1$ picking it are small. Thus, it does not affect the mean rewards of the experts much. This leads to our Modified methods achieving constant regret. However, in Setting 2, this arm is associated with high rewards, leading to the high uncertainty. In this case, the performance of the modified D-UCB algorithm is comparable to that of vanilla UCB, while modified ED-UCB is slightly worse. We note that ED-UCB is subpar in this case as it works off of inaccurate information about the environment (specifically, its value of $p_V$).

\end{document}